%% file: main.tex
\PassOptionsToPackage{table}{xcolor}
\documentclass{article}

\PassOptionsToPackage{numbers, compress}{natbib}
\usepackage[main, final]{neurips_2026}
\usepackage{enumitem}
\usepackage{bm}
\newcommand{\mW}{\bm{W}}
\usepackage{algorithm}
\usepackage{algpseudocode}
\usepackage{wrapfig}
\usepackage{tabularx}
\usepackage{array}

\usepackage[utf8]{inputenc}
\usepackage[T1]{fontenc}
\usepackage{fontawesome5}
\usepackage{hyperref}
\usepackage{url}
\usepackage{booktabs}
\usepackage{amsfonts}
\usepackage{nicefrac}
\usepackage{microtype}
\usepackage{xcolor}

\usepackage{tikz}
\usetikzlibrary{positioning, arrows.meta, shapes.geometric, calc}

\title{Rethinking Layer-wise Model Merging \\ through Chain of Merges}

\author{%
  Pietro Buzzega \hspace{2em}Riccardo Salami \hspace{2em}Angelo Porrello \hspace{2em}Simone Calderara\\ \\
  AImageLab - University of Modena and Reggio Emilia, Modena, Italy \\
  \texttt{name.surname@unimore.it} \\
}

\input{macro}
\input{math_commands}

\begin{document}

\maketitle

\input{chapters/00_abstract}
\input{chapters/1_introduction}
\input{chapters/2_background}
\input{chapters/3_method}
\input{chapters/4_experiments}
\input{chapters/5_analysis}
\input{chapters/6_related}
\input{chapters/7_conclusions}

\begin{ack}
We acknowledge the CINECA award under the ISCRA initiative, for the availability of high-performance computing resources and support.
This work is supported by the Horizon Europe Chips Joint Undertaking under the NexTArc project (HORIZON-JU-Chips-2024-2-RIA).
NexTArc -- Next Generation Open Innovations in Trustworthy Embedded AI Architectures for Smart Cities, Mobility and Logistics (Grant Agreement ID: 101194287, DOI: 10.3030/101194287).
Additionally, the research activities of Angelo Porrello have been partially supported by the Department of Engineering ``Enzo Ferrari'' through the program FAR2025DIP (CUP E93C25000370005).
We also gratefully acknowledge Tetra Pak for funding the PhD position of Riccardo Salami.
\end{ack}

\bibliography{bibliography}
\bibliographystyle{plainnat}

\newpage
\appendix
\crefalias{section}{appendix}
\crefalias{subsection}{subappendix}
\crefalias{subsubsection}{subsubappendix}
\onecolumn
\input{supplementary/0_supplementary}

\end{document}

%% file: macro.tex
\usepackage{xspace}
\usepackage{graphicx}
\usepackage{amssymb}
\usepackage{amsmath}
\usepackage{dsfont}
\usepackage{multirow}
\usepackage{float}
\usepackage{lipsum}
\usepackage{nicefrac}
\usepackage{circledsteps}
\usepackage{enumitem}
\usepackage{colortbl}
\usepackage{graphicx}
\usepackage{booktabs}
\usepackage{siunitx}
\usepackage{breqn}
\usepackage{lipsum}
\usepackage{subcaption}
\usepackage[capitalize,noabbrev]{cleveref}
\usepackage{duckuments}
\usepackage{tikzducks}
\usepackage{tablefootnote}
\usepackage{pifont}
\usepackage{makecell}
\usepackage{tikz}
\usepackage{tcolorbox}
\usepackage{wrapfig}
\usepackage{needspace}

\definecolor{lightgray}{gray}{0.95}
\definecolor{midgray}{gray}{0.55}
\definecolor{steelblue}{HTML}{4D82B7}
\definecolor{davysgrey}{rgb}{0.33, 0.33, 0.33}
\definecolor{LightCyan}{rgb}{0.88,1,1}
\definecolor{ao(english)}{rgb}{0.0, 0.5, 0.0}

\usepackage[first=0, last=9]{lcg}

\newcommand{\cmark}{\ding{51}}%
\newcommand{\xmark}{\ding{55}}%

\newcommand{\Star}[1]{#1\ensuremath{^*}\kern-\scriptspace}

\creflabelformat{equation}{#2\textup{#1}#3}
\crefname{section}{Sec.}{Secs.}
\crefname{appendix}{App.}{Apps.}
\crefname{subappendix}{App.}{Apps.}
\crefname{subsubappendix}{App.}{Apps.}
\crefname{table}{Tab.}{Tabs.}
\crefname{figure}{Fig.}{Figs.}
\crefname{equation}{Eq.}{Eqs.}
\crefname{algorithm}{Alg.}{Algs.}

\newcommand{\methname}{CoM\xspace}

\makeatletter
\DeclareRobustCommand\onedot{\futurelet\@let@token\@onedot}
\def\@onedot{\ifx\@let@token.\else.\null\fi\xspace}

\def\eg{\emph{e.g}\onedot}

\DeclareMathOperator*{\minimize}{minimize}

\definecolor{algc1}{HTML}{f7d779}
\definecolor{algc2}{HTML}{9fc5fc}

\usepackage{array}
\newcommand{\PreserveBackslash}[1]{\let\temp=\\#1\let\\=\temp}
\newcolumntype{C}[1]{>{\PreserveBackslash\centering}p{#1}}
\newcolumntype{R}[1]{>{\PreserveBackslash\raggedleft}p{#1}}
\newcolumntype{L}[1]{>{\PreserveBackslash\raggedright}p{#1}}

\newcommand{\Ttop}{\mkern-0.5mu\raisebox{0.55ex}{$\scriptstyle \top$}}

%% file: math_commands.tex
\usepackage{amsmath,amsfonts,bm}

\def\eqref#1{equation~\ref{#1}}

\def\1{\bm{1}}

\def\vx{{\bm{x}}}

\def\mG{{\bm{G}}}

\def\mW{{\bm{W}}}
\def\mX{{\bm{X}}}

\DeclareMathAlphabet{\mathsfit}{\encodingdefault}{\sfdefault}{m}{sl}
\SetMathAlphabet{\mathsfit}{bold}{\encodingdefault}{\sfdefault}{bx}{n}

\DeclareMathOperator*{\argmin}{arg\,min}

%% file: chapters/00_abstract.tex
\begin{abstract}
Model merging has emerged as a simple yet effective method to fuse models without re-training. Existing techniques operate at the level of individual layers, thereby overlooking the inter-layer dependencies inherent in deep networks. We show that this simplification induces distributional mismatches in intermediate activations during merging, as changes applied to early layers fail to propagate to downstream ones. We identify these mismatches as a form of \textit{covariate shift} that compounds across the network. To address this issue, we propose \textit{Chain of Merges} (CoM), a novel procedure that sequentially merges weights across layers while accounting for inter-layer interactions. In particular, CoM mitigates covariate shift through a series of regression problems, where input activations are recomputed at each step to reflect the updated representations. Additionally, we introduce a dynamic weighting mechanism that prioritizes task-layer pairs most susceptible to performance degradation, improving robustness during merging. Experiments on standard benchmarks demonstrate that CoM achieves state-of-the-art performance across heterogeneous tasks. $~~~~~~${\hypersetup{pdfborder={0 0 0}}\href{https://github.com/aimagelab/chain-of-merges}{\textcolor{blue!65!black}{\faGithub\,$~$\texttt{github.com/aimagelab/chain-of-merges}}}}\hspace{-0.25em}
\end{abstract}

%% file: chapters/1_introduction.tex
\section{Introduction}
The availability of large-scale pretrained models has reshaped machine learning~\cite{radford2021learning, touvron2023llama}, with fine-tuning offering an accessible path to state-of-the-art performance across domains~\cite{raffel2020exploring, wang2018glue}. As these models are adapted to specialized tasks, a natural question arises: \textit{how can we combine task-specific checkpoints without retraining?} This challenge, known as model merging~\cite{ilharco2023editing, matena2022merging, wortsman2022model}, enables modularity, knowledge reuse, and efficient deployment.

Since specialized modules are trained independently, there is no guarantee that their weights can be seamlessly combined~\cite{yadav2023ties,stoica2023zipit}. Naive strategies like weight averaging~\cite{mcmahan2017communication,wortsman2022model} often cause performance degradation~\cite{tang2024merging,daheim2024model,tam2023merging}. To address this, current literature relies on a variety of heuristics, \eg interference mitigation~\cite{yadav2023ties,yu2024language}, permutation-based parameter alignment~\cite{ainsworth2023git,singh2020model}, critical weight preservation~\cite{matena2022merging,lee2025dynamic}, and interpolation within orthonormal or task-adaptive subspaces~\cite{marczak2025no,gargiulo2025task,tam2023merging}.

While these approaches achieve competitive performance, they still present notable limitations. They often rely on problem-specific design choices and extensive hyperparameter tuning, while lacking a unifying theoretical foundation that provides strong guarantees in real-world deployment settings. Motivated by these shortcomings, a complementary line of work focuses on \textbf{aligning model activations}~\cite{stoica2023zipit,jin2023dataless,tatro2020optimizing,jordan2022repair}, among which RegMean~\cite{jin2023dataless} is the closest to our work. By operating in activation space, these methods cast model merging as an interpretable, layer-wise optimization problem aimed at preserving the functional behavior of each layer after merging. In many cases, these are simple regression problems that admit closed-form solutions, leading to more stable behavior. This perspective generally provides stronger theoretical grounding, albeit at the cost of requiring additional data to estimate activation statistics.

While these activation-based approaches lay the foundation for more principled, interpretable, and trustworthy model merging, they overlook a \textbf{simple but crucial} factor: each layer is conditioned on the outputs of preceding computations. Indeed, these approaches merge layers \textbf{independently}, proceeding sequentially from one layer to the next. As this process unfolds, the output distribution of each layer shifts due to the merging operations applied to preceding layers. Current methods, starting from RegMean~\cite{jin2023dataless}, fail to account for this shift and instead rely on the original activation distributions -- namely, those computed from the individual models prior to merging.

As a {\setlength{\fboxsep}{1.5pt}\colorbox{teal!30}{principal contribution}}, we identify this issue as a form of \textit{covariate shift}~\cite{ioffe2015batch}, a well-known phenomenon in training dynamics where rapidly changing early-layer activations induce unstable output distributions that hinder downstream learning~\cite{arpit2016normalization}. By analogy, we refer to its manifestation in model merging as \textit{\textbf{Merging Covariate Shift}} (MCS), which arises when early layers are modified through merging, causing shifts in the inputs to subsequent layers. In this respect, we show that merging layers in isolation inherently alters the distribution of their output activations, resulting in misaligned inputs for downstream layers. This effect compounds across the network: even small discrepancies propagate and amplify, causing progressively larger mismatches and consequent performance degradation. This naturally leads to the following research question:
\vspace{-0.1em}
\begin{tcolorbox}[
    colback=lightgray!35,
    colframe=black!45,
    boxrule=0.3pt,
    arc=3mm,
    auto outer arc,
    width=0.95\linewidth,
    center,
    halign=center,
    boxsep=3pt,
    left=2pt, right=2pt, top=3pt, bottom=3pt
]
\emph{How can we design a layer-wise merging procedure that accounts for inter-layer interactions and prevents the propagation of distributional mismatches across layers?}
\end{tcolorbox}
\vspace{-0.1em}

\input{figures/intro_fig}
Building on this crucial yet overlooked limitation, we propose \textbf{Chain of Merges (CoM)}, illustrated in \cref{fig:mcs_qualitative}. CoM is a recursive, layer-wise merging method that updates activation statistics after each merging step, replacing the original task-specific activations with those produced by the partially merged model. In doing so, CoM explicitly accounts for inter-layer dependencies and the MCS issue. As a {\setlength{\fboxsep}{1.5pt}\colorbox{orange!25}{second result}}, we show that this simple modification can dramatically improve performance, enabling our approach to surpass the state of the art by a significant margin.

As a {\setlength{\fboxsep}{1.5pt}\colorbox{purple!25}{{third contribution}}}, we emphasize that not all tasks are equally important or equally difficult. An activation-based approach should account for this heterogeneity by prioritizing certain tasks over others during optimization. In particular, we hypothesize that models which deviate more strongly from the base model are also more susceptible to performance degradation. We introduce a \textbf{dynamic weighting mechanism} into our framework, designed to better preserve these more vulnerable representations. We further show that a proxy for such susceptibility -- and the resulting performance degradation -- can be directly estimated from the feature correlation matrix, which is the same input we leverage to solve the optimization problem underlying CoM.

To sum up:
{\setlength{\fboxsep}{1pt}\colorbox{teal!30}{\textbf{(i)}}} we identify and analyze covariate shift in model merging, showing that activation mismatches accumulate in deep networks, and propose Chain of Merges (\methname), a method that progressively merges parameters while updating activation statistics to ensure consistency across the merged network;
\setlength{\fboxsep}{1pt}\colorbox{purple!25}{\textbf{(ii)}} we introduce an importance-weighting mechanism that biases the solution toward task-layer pairs most divergent from the base model, thereby preserving their features; and
\setlength{\fboxsep}{1pt}\colorbox{orange!25}{\textbf{(iii)}} we evaluate \methname on standard model merging benchmarks across language and vision settings, showing that it outperforms existing methods by a large margin on LoRA fine-tuning.

%% file: figures/intro_fig.tex
\begin{figure}[t]
    \centering
    \resizebox{0.95\linewidth}{!}{
    \begin{tikzpicture}[
        >=stealth,
        block/.style={rectangle, draw, rounded corners, minimum width=2.4cm, minimum height=0.9cm, align=center, font=\sffamily\small},
        modelA/.style={fill=blue!10, draw=blue!70!black, thick},
        modelB/.style={fill=orange!10, draw=orange!70!black, thick},
        merged/.style={fill=teal!20, draw=teal!80!black, thick},
        data/.style={ellipse, draw=gray!70, fill=gray!5, font=\sffamily\small, inner sep=2pt},
        merge_flow/.style={->, thick, draw=gray!80},
        merge_param/.style={->, thick, dashed, draw=gray!80},
        infer_flow/.style={->, thick, draw=red!80},
        title/.style={font=\sffamily\large\bfseries, align=center}
    ]

    \node[title] (TitleA) at (0,0) {(a) Independent Merging (Current)};

    \node[block, modelA] (L1A_A) at (-3.2, -1.1) {Layer $l$\\Model A};
    \node[block, modelB] (L1B_A) at (3.2, -1.1) {Layer $l$\\Model B};
    \node[block, merged] (L1M_A) at (0, -1.1) {Layer $l$\\Merged Model};

    \draw[merge_param] (L1A_A) -- (L1M_A);
    \draw[merge_param] (L1B_A) -- (L1M_A);

    \node[data] (ActA_A) at (-3.2, -2.8) {Acts $\mX_A^l$};
    \node[data] (ActB_A) at (3.2, -2.8) {Acts $\mX_B^l$};
    \node[data, draw=red!80, fill=red!10, font=\sffamily\small\bfseries] (ActM_A) at (0, -2.8) {Acts $\widehat{\mX}^l$};

    \draw[merge_flow] (L1A_A) -- (ActA_A);
    \draw[merge_flow] (L1B_A) -- (ActB_A);
    \draw[infer_flow] (L1M_A) -- (ActM_A);

    \node[block, modelA] (L2A_A) at (-3.2, -5.2) {Layer $l+1$\\Model A};
    \node[block, modelB] (L2B_A) at (3.2, -5.2) {Layer $l+1$\\Model B};
    \node[block, merged] (L2M_A) at (0, -5.2) {Layer $l+1$\\Merged Model};

    \draw[merge_param] (ActA_A) -- (L2A_A) node[midway, left, align=right, font=\footnotesize, text=gray!80!black, xshift=-0.1cm] {Stats \\computed\\on $\mX_A^l$};
    \draw[merge_param] (ActB_A) -- (L2B_A) node[midway, right, align=left, font=\footnotesize, text=gray!80!black, xshift=0.1cm] {Stats \\computed\\on $\mX_B^l$};

    \draw[infer_flow] (ActM_A) -- (L2M_A) node[midway, left, text=red!80!black, font=\sffamily\scriptsize\bfseries, align=right, xshift=-0.1cm] {Mismatch!\\Shifted input ($\widehat{\mX}^l$)};

    \draw[<->, red, thick, dashed, bend right=15] (ActA_A) to node[midway, above, font=\sffamily\scriptsize\bfseries] {MCS} (ActM_A);
    \draw[<->, red, thick, dashed, bend left=15] (ActB_A) to node[midway, above, font=\sffamily\scriptsize\bfseries] {MCS} (ActM_A);

    \begin{scope}[shift={(1.4, -4.2)}, scale=0.65]
        \draw[->, thick, gray] (-1.2, 0) -- (1.8, 0);
        \draw[thick, gray, fill=gray!20, opacity=0.7] (-0.8,0) .. controls (-0.5,0) and (-0.3,1.2) .. (0,1.2) .. controls (0.3,1.2) and (0.5,0) .. (0.8,0);
        \draw[thick, red!80, dashed, fill=red!10, opacity=0.8] (-0.1,0) .. controls (0.2,0) and (0.4,1) .. (0.7,1) .. controls (1.0,1) and (1.2,0) .. (1.5,0);
        \node[font=\sffamily\scriptsize, text=gray!80!black, anchor=east] at (-0.2, 1.0) {Expected};
        \node[font=\sffamily\scriptsize, text=red!80, anchor=west] at (0.85, 0.8) {Shifted};
    \end{scope}

    \draw[merge_param] (L2A_A) -- (L2M_A);
    \draw[merge_param] (L2B_A) -- (L2M_A);

    \begin{scope}[xshift=10cm]
    \node[title] (TitleB) at (0,0) {(b) Chain of Merges (CoM)};

    \node[block, modelA] (L1A_B) at (-3.2, -1.1) {Layer $l$\\Model A};
    \node[block, modelB] (L1B_B) at (3.2, -1.1) {Layer $l$\\Model B};
    \node[block, merged] (L1M_B) at (0, -1.1) {Layer $l$\\Merged Model};

    \draw[merge_param] (L1A_B) -- (L1M_B);
    \draw[merge_param] (L1B_B) -- (L1M_B);

    \node[data, draw=green!60!black, fill=green!10, font=\sffamily\small\bfseries] (ActM_B) at (0, -2.8) {Updated Acts $\widehat{\mX}^l$};
    \draw[infer_flow] (L1M_B) -- (ActM_B);

    \node[block, modelA] (L2A_B) at (-3.2, -5.2) {Layer $l+1$\\Model A};
    \node[block, modelB] (L2B_B) at (3.2, -5.2) {Layer $l+1$\\Model B};
    \node[block, merged] (L2M_B) at (0, -5.2) {Layer $l+1$\\Merged Model};

    \draw[infer_flow] (ActM_B) -- (L2M_B) node[midway, left, text=green!60!black, font=\sffamily\scriptsize\bfseries, align=right, xshift=0.05cm] {Consistent\\input};

    \draw[merge_param, bend right=15] (ActM_B) to node[midway, left, align=right, font=\footnotesize, text=gray!80!black, xshift=-0.4cm] {Stats updated\\via $\widehat{\mX}^l$} (L2A_B);
    \draw[merge_param, bend left=15] (ActM_B) to node[midway, right, align=left, font=\footnotesize, text=gray!80!black, xshift=0.4cm] {Stats updated\\via $\widehat{\mX}^l$} (L2B_B);

    \begin{scope}[shift={(1.0, -4.2)}, scale=0.65]
        \draw[->, thick, gray] (-1.2, 0) -- (1.8, 0);
        \draw[thick, gray, fill=gray!20, opacity=0.7] (-0.8,0) .. controls (-0.5,0) and (-0.3,1.2) .. (0,1.2) .. controls (0.3,1.2) and (0.5,0) .. (0.8,0);
        \draw[thick, green!60!black, dashed, fill=green!10, opacity=0.6] (-0.8,0) .. controls (-0.5,0) and (-0.3,1.2) .. (0,1.2) .. controls (0.3,1.2) and (0.5,0) .. (0.8,0);
        \node[font=\sffamily\footnotesize, text=green!60!black, anchor=north] at (0, -0.1) {Aligned};
    \end{scope}

    \draw[merge_param] (L2A_B) -- (L2M_B);
    \draw[merge_param] (L2B_B) -- (L2M_B);
    \end{scope}

    \draw[loosely dashed, gray, thick] (5.0, 0.5) -- (5.0, -5.8);

    \begin{scope}[shift={(5.0, -6.2)}]
        \node[rectangle, draw=gray!40, fill=gray!5, rounded corners, minimum width=13cm, minimum height=0.6cm] (Legend) at (0,0) {};

        \draw[merge_flow] (-5.4, 0) -- (-4.8, 0);
        \node[right, font=\sffamily\footnotesize, text=black!80] at (-4.7, 0) {Statistics from individual models};

        \draw[infer_flow] (0.5, 0) -- (1.1, 0);
        \node[right, font=\sffamily\footnotesize, text=black!80] at (1.2, 0) {Statistics from partially-merged model};
    \end{scope}

    \end{tikzpicture}
    }
    \caption{(a) Independent merging induces a distribution shift as Layer $l+1$ receives mismatched activations. (b) CoM recomputes statistics using updated activations, realigning the distribution.}
    \label{fig:mcs_qualitative}
\end{figure}

%% file: chapters/2_background.tex
\section{Towards Understanding \textit{Merging Covariate Shift}}
\paragraph{Background.} Model merging aims to combine a collection of $N$ models, all sharing an identical architecture, independently trained on distinct input datasets. Each model comprises $L$ linear layers, which are the target of the merging procedure. For a given layer $l \in \{1, 2, \dots, L \}$, the set of corresponding weight matrices is denoted as $\{\mW_i^l\}_{i=1}^{N}$, all having the same dimensions. Notably, our study focuses on Transformer-based architectures, where linear projections constitute the vast majority of the parameter count (approximately 95\%).

Among activation-based approaches, RegMean~\cite{jin2023dataless} proposes to find a single linear transformation, $\mW_M^l$, that best approximates the behavior of the original layers when applied to their respective inputs $\{\bm{X}_i^l\}_{i=1}^{N}$, with $\mX_i^l \in \mathbb{R}^{d \times T_i}$ having one column per input token. This is accomplished by minimizing the following objective:
\begin{equation}
\label{eq:omega}
\minimize ~ \Omega^l = \sum_{i=1}^{N}{\left\|\mW^l_M\mX_i^l - \mW_i^l\mX_i^l\right\|^2_{F}}.
\end{equation}
By differentiating $\Omega^l$ with respect to $\mW_M^l$ and setting the result to zero, we can obtain a closed-form solution for the optimal merged layer:
\begin{equation}
\label{eq:regmean}
\mW_M^l = \left(\sum_{i = 1}^N \mW_i^l \mX_i^l {\mX_i^l}^\top\right) \left(\sum_{i = 1}^N \mX_i^l {\mX_i^l}^\top\right)^{-1}.
\end{equation}
Here, the Gram matrix $\mX_i^l {\mX_i^l}^\top \in \mathbb{R}^{d \times d}$ captures the pairwise correlations between input features.

\subsection{Merging Covariate Shift}
\label{sec:model}

To solve the problem in \cref{eq:omega}, RegMean~\cite{jin2023dataless} and related methods compute layer-wise statistics of the network, using as reference the model trained individually on each task. In particular, denoting by $\mX_i^{l}$ the activations of model $i$ at layer $l$, these are defined recursively as
\begin{equation}
\mX_i^{l+1} = \boldsymbol{\sigma}\!\left( \mW_i^{l} \mX_i^{l} \right), \quad l = 1, \dots, L-1,
\end{equation}
starting from the input $\mX_i^1$. For simplicity of notation, we assume that the activation function $\boldsymbol{\sigma}$ is the same across all layers. Consequently, up to layer $l+1$, for each model $i$ we obtain the following \textbf{target activations} for the RegMean problem:
\begin{equation}
\mX_i^{2} = \boldsymbol{\sigma}\!\left( \textcolor{red!75!black}{\mW_i^{1}} \mX_i^{1} \right),
\dots,
\mX_i^l = \boldsymbol{\sigma}\!\left( \textcolor{red!75!black}{\mW_i^{l-1}} \mX_i^{l-1} \right),
\mX_i^{l+1} = \boldsymbol{\sigma}\!\left( \textcolor{red!75!black}{\mW_i^{l}} \mX_i^{l} \right).
\end{equation}
At inference time, the original parameters $\mW_i^{l-1}$ are replaced with their merged counterparts $\mW_M^{l-1}$, obtained by solving \cref{eq:regmean}. Consequently, the activations shift accordingly and become:
\begin{equation}
\label{eq:activation_layer}
\widehat{\mX}_i^{2} = \boldsymbol{\sigma}\!\left( \textcolor{blue}{\mW_M^{1}} \mX_i^{1} \right),
\dots,
{\widehat{\mX}}_i^l = \boldsymbol{\sigma}\!\left( \textcolor{blue}{\mW_M^{l-1}} {\widehat{\mX}}_i^{l-1} \right),
{\widehat{\mX}}_i^{l+1} = \boldsymbol{\sigma}\!\left( \textcolor{blue}{\mW_M^{l}} {\widehat{\mX}}_i^{l} \right).
\end{equation}
\textbf{Problem.} The statistics used to merge layer $l+1$ are computed using the original parameters $\textcolor{red!75!black}{\mW_i^1, \dots, \mW_i^{l}}$ prior to merging. However, at inference time, the preceding layers have already been replaced by their merged counterparts $\textcolor{blue}{\mW_M^1, \dots, \mW_M^{l}}$. As a result, the input activations to layer $l+1$ differ from those used during the estimation of the merging statistics. This mismatch induces a shift in the activation distributions, analogous to internal covariate shift, which we refer to as \textit{Merging Covariate Shift} (MCS).

To make MCS measurable, we follow the common view of covariate shift as a distributional discrepancy~\citep{huang2020internal}, and model the activation inputs of each layer as multivariate Gaussians. In particular, we consider the pre-merging activations $\mX_i^l$, computed using the original parameters $\textcolor{red!75!black}{\mW_i^{1:l-1}}$, and the post-merging activations ${\widehat{\mX}}_i^l$, obtained with the merged counterparts $\textcolor{blue}{\mW_M^{1:l-1}}$. We hence quantify the distance between these two distributions using the squared 2-Wasserstein distance (also known as Fr\'echet distance)~\citep{villani2008optimal,dowson1982frechet}, which admits a closed form for Gaussians:
\begin{equation}
\begin{aligned}
\label{eq:mcs}
    \operatorname{MCS}^l &= \sum_{i=1}^{N}~\left[~\left\|\mu_i^l - \widehat\mu_i^l\right\|_2^2 +
\operatorname{Tr}\Big[ \Sigma_i^l + \widehat\Sigma_i^l - 2\left( (\widehat\Sigma_i^l)^{\nicefrac{1}{2}}\, \Sigma_i^l \, (\widehat\Sigma_i^l)^{\nicefrac{1}{2}} \right)^{\nicefrac{1}{2}} \Big]\right],
\end{aligned}
\end{equation}
where $\mu_i^l,\Sigma_i^l$ are the mean and covariance of $\mX_i^l$, and $\widehat\mu_i^l,\widehat\Sigma_i^l$ are those of $\widehat{\mX}_i^l$.

\input{figures/fig_mcs}
As shown in \cref{fig:mcs}, evaluated on 8-Vision~\cite{ilharco2022patching}, MCS is present across all layers and increases sharply with depth, suggesting that mismatches compound as they propagate through the network. Notably, this effect is significantly more pronounced in larger models (ViT-L), suggesting that model scale exacerbates the accumulation of distributional mismatches. We also analyze the two projections before and after attention separately, as they exhibit different magnitudes: QKV projections show substantially higher MCS than output projections, indicating that certain components of the attention mechanism are more sensitive to merging-induced perturbations.

%% file: figures/fig_mcs.tex
\setlength{\intextsep}{0pt}

\begin{wrapfigure}[13]{r}{0.45\linewidth}
\centering
\includegraphics[width=\linewidth]{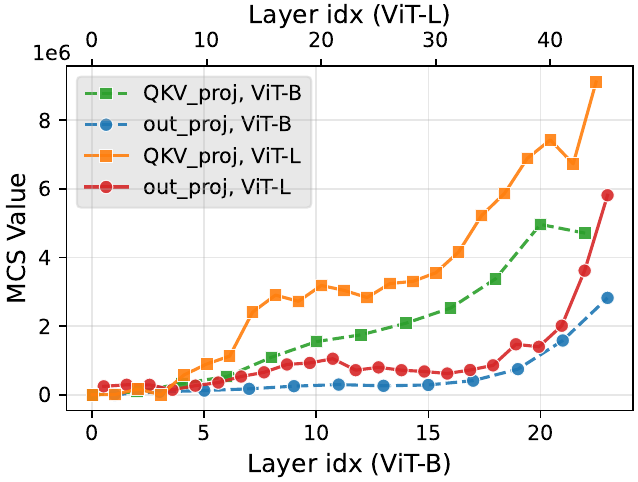}
\caption{MCS across layers.}
\label{fig:mcs}

\end{wrapfigure}

%% file: chapters/3_method.tex
\section{Chain of Merges}
\textbf{Recursive Dependence.}
To address MCS, we revise the closed-form solution presented in \cref{eq:regmean}. Instead of relying on the stale inputs $\mX_i^{l}$, computed using the original parameters $\textcolor{red!75!black}{\mW_i^1, \dots, \mW_i^{l-1}}$, we employ the activations $\widehat{\mX}_i^{l}$ produced by the preceding \textbf{merged} layers, i.e., using $\textcolor{blue}{\mW_M^1, \dots, \mW_M^{l-1}}$. These correspond to the actual inputs received by layer $l$ at inference time, once all preceding layers have been merged. Substituting $\mX_i^l$ with $\widehat{\mX}_i^l$ yields the revised expression for the merged weights:
\begin{equation}
\label{eq:com}
\mW_M^l = \left(\sum_{i = 1}^N \mW_i^l \widehat{\mX}_i^l\widehat{\mX}_i^{l\vspace{-1cm}\Ttop}\right) \left(\sum_{i = 1}^N \widehat{\mX}_i^l\widehat{\mX}_i^{l\vspace{-1cm}\Ttop}\right)^{-1}.
\end{equation}
This substitution induces a \textit{recursive} dependence: the inputs to layer $l$ now depend on the outputs of layer $(l-1)$, which are themselves computed using the merged weights $\textcolor{blue}{\mW_{M}^{l-1}}$ and the inputs to layer $(l-1)$. In turn, these depend on the merged outputs of layer $(l-2)$, and so forth. Hence, at every preceding layer all unmerged activations and parameters must be replaced by their merged counterparts, propagating the correction through the entire network. By ensuring that the statistics computed at each stage reflect the accumulated effect of all preceding merges, this autoregressive scheme mitigates Merging Covariate Shift by design.

We can summarize the CoM procedure as follows:
\begin{tcolorbox}[
    colback=gray!10,
    colframe=black!40,
    boxrule=0.3pt,
    arc=2mm,
    boxsep=3pt,
    left=4pt, right=4pt, top=4pt, bottom=4pt
]

\textbf{Initial step.} The recursive chain starts at the first layer, where the inputs correspond to raw data and are unaffected by prior merging. Thus, the merged weights can be computed using \cref{eq:regmean}.

\smallskip
\textbf{Recursive step.}
The merged weights obtained at the initial step are used to propagate consistent activations throughout the network. For subsequent layers $l = 2, \dots, L$, the algorithm proceeds iteratively, alternating between computing the activations $\widehat{\mX}_i^{\,l}$ -- which account for merging in preceding layers -- and updating the merged weights $\textcolor{blue}{\mW_M^{l}}$ according to \cref{eq:com}.
\end{tcolorbox}

\smallskip
\textbf{Fast implementation.} Importantly, our recursive formulation incurs a computational cost that is \textbf{linear} in the number of merged models. The overall forward computation requires exactly one forward pass per model -- matching the cost of RegMean -- despite the sequential nature of the merging procedure and the fact that activation statistics are updated at every layer. To achieve this efficiency, we implement CoM by interleaving merging and propagation at the layer level. Specifically, we leverage forward pre-hooks attached to each fine-tuned linear projection and maintain, for each task $i$, a single activation buffer that stores the input to the \textbf{next} target layer in the partially merged model.
\subsection{Correlation-based importance}
\label{sec:importance}
Although our method addresses the Merging Covariate Shift, the merged layer still cannot exactly reproduce the outputs of every task-specific layer. A single matrix $\mW_M^l$ must match the activations $\mW_i^l\widehat{\mX}_i^l$ of $N$ distinct layers on their respective inputs, and, in general, no single linear map achieves this for all tasks at once. The least-squares solution of \cref{eq:com} therefore leaves a residual error, which perturbs the activations of the merged model on every task.

Given that an error during merging is therefore unavoidable, the remaining question is how to distribute this error across the different tasks. \textit{Are there tasks that are more important than others?}

\smallskip
\textbf{Task Importance.} The goal of model merging is to retain performance across all tasks; however, their importance is not uniform. In this regard, we distinguish two categories of tasks:
\begin{itemize}
    \item \texttt{Category (a)} Tasks that are semantically \textbf{similar} to the pretraining distribution, and therefore naturally benefit from the representations of the base model.
    \item \texttt{Category (b)} Tasks that are semantically \textbf{distant} from pretraining, which cannot rely on pretrained features and are thus more prone to performance degradation after merging.
\end{itemize}

We argue that preserving \texttt{Category (b)} tasks is more critical. Indeed, as they deviate more from the base model, they are also more likely to suffer larger losses if their task-specific weights are not adequately incorporated into the merged model. To account for this \textbf{asymmetry}, we assign each task-layer pair an importance weight $\omega^{\,l}_i$, reflecting the categorization introduced above: \textit{the more} distant a task is from the base model, \textit{the more} important it is to preserve it.

\smallskip
\textbf{Feature correlation as a proxy.} We require a task-layer importance score $\omega^l_i$ that can be computed during merging, without labels. We construct this score based on the \textbf{redundancy} of input activations:

\begin{tcolorbox}[
    colback=lightgray!35,
    colframe=black!45,
    boxrule=0.3pt,
    arc=3mm,
    auto outer arc,
    width=0.95\linewidth,
    center,
    halign=center,
    boxsep=3pt,
    left=2pt, right=2pt, top=3pt, bottom=3pt
]
If, for a given task, the activations at a certain layer are highly redundant, we interpret this as a signal that the model \textbf{has collapsed} its representations to accommodate the task, and thus that the task \textbf{has significantly deviated} from the base model.
\end{tcolorbox}

For an empirical assessment of this intuition, in \cref{sec:feature_correlation} (\cref{fig:importance}) we show that weight scores based on activation redundancy are \textit{linearly correlated} with a performance-based notion of task importance.

More precisely, we measure how strongly different activation directions co-vary, and use this as a signal of how much task $i$ has specialized away from the pretrained features at layer $l$. Large-scale pretraining encourages representations that are broad and approximately decorrelated, as they must support many downstream uses~\cite{cogswell2015reducing,morcos2018importance}. When a model is fine-tuned, this generality is partially sacrificed: activations collapse onto a narrower subspace, and previously independent feature directions begin to align. The amount of off-diagonal mass in the activation correlation matrix thus provides a direct measure of how far the task has pushed the layer away from the pretrained, decorrelated state. Tasks that are semantically distant from pretraining rely most on this specialization, and are therefore assigned greater weight during optimization.

CoM already produces the inputs $\widehat{\mX}^l_i$ required by \cref{eq:activation_layer}. We reuse them after normalizing each token (column of $\widehat{\mX}^l_i$) to unit $\ell_2$ norm (see \cref{app:further_imp_details} for more details) to obtain $\tilde{\mX}^l_i$, and form the correlation matrix $\mG^{\,l}_i = \frac{1}{T_i}\tilde{\mX}_i^l \tilde{\mX}_i^{l\Ttop}$. The importance score is the off-diagonal $\ell_1$ mass of $\mG^l_i$, which captures feature coupling:
\begin{equation}
\omega^l_i = \sum_{p<q} \big|(\mG^l_i)_{pq}\big|.
\end{equation}
A large $\omega^l_i$ flags a layer that has narrowed toward a task-specific subspace and is therefore fragile under merging, while a small value indicates near-orthogonal features and low sensitivity. In \cref{app:appendix_task_similarity}, we further relate the decorrelation of pretrained features to generalization, and show that computing $\omega_i^l$ with the merged model instead of the pretrained one yields closely matching scores.

\smallskip
\textbf{Importance-weighted merging.} To incorporate task- and layer-specific importance, we extend our objective $\Omega$ with the weighting factor $\omega^{\,l}_i$, biasing the model toward more sensitive tasks. The resulting merging rule for $\mW_M^{\,l}$ becomes:
\begin{equation}
\begin{aligned}
\label{eq:cm_weighted}
&\argmin_{\mW_M^{\,l}} ~ \sum_{i=1}^{N}{\frac{\omega_{i}^{\,l}}{T_i}\left\|\mW^l_M\tilde{\mX}_i^l - \mW_i^l\tilde{\mX}_i^l\right\|^2_{F}} \;=\;
\Bigl(\sum_{i=1}^{N} \omega_{i}^{\,l}\,
       \mW_{i}^{\,l}\,\mG_i^{\,l}\Bigr)
\Bigl(\sum_{i=1}^{N} \omega_{i}^{\,l}\,
       \mG_i^{\,l}\Bigr)^{-1}.
\end{aligned}
\end{equation}
The complete procedure is summarized in \cref{alg:chain_merges}, with further implementation details in \cref{app:further_imp_details}. The full derivation of \cref{eq:cm_weighted} is provided in \cref{app:derivation_weighted}, while the special case of normalization layers is addressed in \cref{app:derivation_layernorm}.

%% file: chapters/4_experiments.tex
\section{Experiments}
\label{sec:experiments}
\textbf{Experimental Setup.}
We evaluate our approach across both \textbf{vision} and \textbf{language} domains.
For vision, we use the benchmark of~\cite{ilharco2023editing}, which involves merging checkpoints from eight classification datasets: Stanford Cars~\cite{krause20133d}, DTD~\cite{cimpoi2014describing}, EuroSAT~\cite{helber2019eurosat}, GTSRB~\cite{stallkamp2011german}, MNIST~\cite{lecun2002gradient}, RESISC45~\cite{cheng2017remote}, SUN397~\cite{xiao2016sun}, and SVHN~\cite{netzer2011reading}.
For language, following~\cite{stoica2025model,panariello2025accurate}, we evaluate on six datasets: SNLI~\cite{bowman2015snli}, MultiNLI~\cite{williams2017broad}, SICK~\cite{marelli2014semeval}, SciTail~\cite{khot2018scitail}, RTE~\cite{wang2018glue}, and QNLI~\cite{wang2018glue}.
SNLI, MultiNLI, and SICK are three-way classification tasks (entailment, contradiction, neutral), while SciTail, RTE, and QNLI are binary classification tasks.

\textbf{Competitors.}
We compare against methods from the main model merging families: task-vector and interference-reduction methods (TA~\cite{ilharco2023editing}, TIES~\cite{yadav2023ties}, DARE~\cite{yu2024language}, Consensus TA~\cite{wang2024localizing}, LiNeS~\cite{wang2025lines}), Fisher or regression-based activation matching (FisherAVG~\cite{matena2022merging}, RegMean~\cite{jin2023dataless}, MaTS~\cite{tam2023merging}), SVD/subspace methods (TSV~\cite{gargiulo2025task}, Iso-C~\cite{marczak2025no}, KnOTS~\cite{stoica2025model}, Core\textsubscript{\tiny TSV}~\cite{panariello2025accurate}, Core\textsubscript{\tiny ISO}~\cite{panariello2025accurate}), and data-driven layer-wise optimization methods (AdaMerging++~\cite{yang2023adamerging}, ProDistill~\cite{xu2025scalable}, Localize-and-Stitch~\cite{he2024localize}).
We refer the reader to \cref{sec:related} for a detailed comparison with these approaches.
Importantly, we compare our approach against merging techniques involving \textbf{both full fine-tuning and LoRA fine-tuning}~\cite{hu2022lora}, evaluating them in two distinct phases.

\textbf{Evaluation protocol.}
All our experiments follow a \textit{static merging} protocol as in~\cite{stoica2025model,ilharco2022patching,gargiulo2025task,panariello2025accurate}: each method outputs a single merged backbone used for all tasks at inference time.
No task identifiers, routing mechanisms, or input-dependent adapters are allowed at test time.
Performance is evaluated using task-specific heads: CLIP zero-shot heads for vision and fine-tuned heads for language.
We report both the average normalized and absolute accuracies of the merged model across all datasets.
\textit{Zero-shot} denotes CLIP's zero-shot performance, while \textit{Individual FT} denotes the performance of each fine-tuned model.

\textbf{Implementation details.}
Following~\cite{stoica2025model,gargiulo2025task,ilharco2023editing}, we use ViT-B/32 and ViT-L/14~\cite{dosovitskiyimage} CLIP encoders as the vision-task backbones.
For natural language tasks, we utilize Llama-3-8B~\cite{grattafiori2024llama}.
Each model is fine-tuned using either LoRA~\cite{hu2022lora} or full fine-tuning.
While the former is applied solely to attention modules (\textit{i.e.}, query, key, value, and output projection layers) with rank $r = 16$, the latter modifies all weights; for all non-linear layers, we employ simple averaging.
To ensure reproducibility and fair comparison, we employ the LoRA fine-tuned checkpoints provided by~\cite{stoica2025model}, and the full fine-tuning checkpoints from~\cite{ilharco2023editing}.
We use \texttt{bfloat16} for NLP tasks -- as it was shown to generally outperform \texttt{float16}~\cite{kalamkar2019study} -- except during Gram-matrix inversion, where \texttt{float32} is used for numerical stability.
To mitigate conditioning issues in language experiments (see \cref{app:further_imp_details}), we use the Moore--Penrose pseudoinverse together with Tikhonov regularization~\cite{hoerl1970ridge} applied as in RegMean~\cite{jin2023dataless} ($\alpha=0.95$ multiplying off-diagonal terms).

\textbf{Calibration and hyperparameter selection.}
For constructing the Gram matrices, we draw balanced examples across tasks and classes; if the number of classes exceeds the number of examples, classes are randomly subsampled.
Calibration examples are drawn from the validation set for CoM (500 examples per task; training set on Llama-3-8B, as RTE and SICK have fewer validation examples), and their labels are only used to balance classes.
CoM results are averaged over three random draws of the calibration set, whereas the other baselines are run once as they are deterministic (either not using data or using the entire training set).
We use the entire training set to estimate RegMean's Gram matrices following the original work~\cite{jin2023dataless}, and discuss the sensitivity of CoM to the construction of the calibration set in \cref{app:calibration_sensitivity}.
Following previous works~\cite{yadav2023ties,stoica2025model,gargiulo2025task}, all hyperparameters are tuned on the entire validation set for each method.
\subsection{Comparison with the State-of-the-Art}
\input{tables/main_results_tables}
\textbf{Vision tasks -- LoRA-based Model Merging.} We refer the reader to \cref{tab:all_average_lora} for a comprehensive comparison. On the smaller ViT-B/32 model, parameter-space baselines retain only moderate performance, with TA, TIES, DARE, Consensus TA, and LiNeS remaining in the mid-$60$s in normalized accuracy. Stronger baselines such as FisherAVG, MaTS, Iso-C, and Core\textsubscript{\tiny ISO} improve over these methods, but CoM achieves a much larger gain, surpassing the best baseline by more than $16$ points (\cref{tab:all_average_lora}). On ViT-L/14, most methods improve and the gap between simple and advanced baselines narrows; nevertheless, CoM remains the best method, reaching $91.35\%$ normalized accuracy versus $86.08\%$ for Core\textsubscript{\tiny ISO}. CoM improves over RegMean by $19.4$ and $8.2$ points on ViT-B/32 and ViT-L/14, respectively, isolating the benefit of correcting MCS within the same activation-matching objective.

\textbf{Language tasks -- LoRA-based Model Merging.} Results on the six natural language benchmarks with Llama-3-8B (\cref{tab:all_average_lora}) show that merging is generally less destructive in this domain, as weight-space baselines such as Task Arithmetic already retain very high normalized accuracy ($90.27\%$ on average). More sophisticated parameter-based techniques provide small gains w.r.t. TA, while RegMean becomes the strongest baseline. Iso-C is a clear outlier, likely because singular-value rescaling interacts poorly with language-model fine-tuning. CoM delivers near-perfect merged performance, surpassing the strongest baseline (RegMean) by $5$ points. We omit FisherAVG and MaTS from this comparison because computing the Fisher Information Matrix or the corresponding linear systems for Llama-3-8B is prohibitively expensive.

\textbf{Vision tasks -- Full Fine-Tuning Model Merging.}
With full fine-tuning, performance improves substantially across all methods\footnote{Core\textsubscript{\tiny TSV} and Core\textsubscript{\tiny ISO} are not applicable to full fine-tuning, as they require factorized (LoRA) updates.} compared to LoRA (\cref{tab:all_average_fft}). On ViT-B/32, simple baselines already become competitive, while ProDistill, TSV, and Iso-C are the strongest baselines, all above $92\%$ normalized accuracy. CoM further advances the state of the art, achieving $94.8\%/86.1\%$ normalized and absolute accuracy. Scaling to ViT-L/14 strengthens most baselines, with AdaMerging++, ProDistill, TSV, and Iso-C all surpassing $95\%$ normalized accuracy. CoM remains on par with the best-performing method, confirming that the proposed recursive activation matching is also effective beyond low-rank adaptation. This overlap occurs solely on ViT-L/14, where performance is essentially saturated (Iso-C: $98.2\%$ \textit{vs.}\ CoM: $97.6\%$ normalized accuracy).

\textbf{Overall.} CoM consistently achieves the best or on-par performance across all settings. The gap with respect to RegMean highlights the importance of mitigating Merging Covariate Shift for effective activation-based merging. Moreover, we highlight that the gains are more pronounced in the LoRA regime than with full fine-tuning.
This is consistent with prior findings that LoRA checkpoints are harder to merge than fully fine-tuned ones~\cite{stoica2025model,zeng2025robustmerge}. We hypothesize that LoRA's restriction of adaptation to a few directions encourages different tasks to encode their knowledge in overlapping dominant directions of the pretrained model. The resulting interference may amplify errors introduced in early layers as they propagate through this restricted adaptation space, making the correction of Merging Covariate Shift particularly important. With full fine-tuning, updates can span broader subspaces, potentially reducing interference; moreover, on ViT-L/14 the benchmark is essentially saturated, leaving little room for improvement (see \cref{app:regime_discussion} for the full discussion).

\textbf{Additional results.} \cref{app:fft_language} reports full fine-tuning results on the language tasks, and \cref{app:more_accuracies} reports the accuracy on each individual dataset for the LoRA-merged models.

%% file: tables/main_results_tables.tex
\begin{table}[t]
\centering
\begin{minipage}[t]{0.57\linewidth}
\centering
\captionof{table}{Normalized and Absolute Average accuracies (\%) of merged LoRA models.}
\label{tab:all_average_lora}
\vspace{0.4em}
\setlength{\tabcolsep}{1.15pt}
\renewcommand{\arraystretch}{1.0}
\rowcolors{6}{lightgray}{}
\begin{tabularx}{\linewidth}{@{} >{\raggedright\arraybackslash\hsize=1.6\hsize}X *{6}{>{\centering\arraybackslash\hsize=0.9\hsize}X} @{}}
\toprule
\rowcolor{white}& \multicolumn{2}{c}{\textbf{ViT-B/32}} & \multicolumn{2}{c}{\textbf{ViT-L/14}} & \multicolumn{2}{c}{\textbf{Llama-3}} \\
\cmidrule(lr){2-3}\cmidrule(lr){4-5}\cmidrule(lr){6-7}
\textbf{Method} & \textbf{Norm} & \textbf{Abs} & \textbf{Norm} & \textbf{Abs} & \textbf{Norm} & \textbf{Abs} \\
\midrule
Zero-shot & 60.6 & 48.1 & 71.4 & 65.3 & 51.1 & 47.4 \\
Indiv. FT & 100.0 & 84.0 & 100.0 & 92.3 & 100.0 & 92.5 \\
\midrule
TA & 63.9 & 51.4 & 74.8 & 68.6 & 90.3 & 83.5 \\
TIES & 64.1 & 51.6 & 75.5 & 69.4 & 90.3 & 83.5 \\
DARE\textsubscript{\tiny TIES} & 64.1 & 51.7 & 75.0 & 68.8 & 90.0 & 83.2 \\
Cons. TA & 65.3 & 52.7 & 76.7 & 70.4 & 90.4 & 83.6 \\
LiNeS & 63.5 & 51.0 & 74.3 & 68.1 & 90.8 & 84.0 \\
FisherAVG & 68.1 & 55.4 & 77.5 & 71.2 & --- & --- \\
RegMean & 73.3 & 61.4 & 83.1 & 76.7 & \underline{94.8} & \underline{87.7} \\
MaTS & 68.3 & 55.6 & 77.5 & 71.1 & --- & --- \\
TSV & 66.8 & 54.1 & 77.8 & 71.5 & 92.2 & 85.2 \\
Iso-C & 73.4 & 59.9 & 84.7 & 77.9 & 70.8 & 65.6 \\
KnOTS\textsubscript{\tiny TIES} & 68.3 & 55.7 & 79.1 & 72.7 & 92.9 & 85.9 \\
CORE\textsubscript{\tiny TSV} & 66.6 & 53.8 & 78.2 & 71.9 & 94.2 & 87.1 \\
CORE\textsubscript{\tiny ISO} & \underline{76.1} & \underline{62.3} & \underline{86.1} & \underline{79.2} & 90.4 & 83.6 \\
\midrule
\textbf{\methname} & \textbf{92.7} & \textbf{78.4} & \textbf{91.4} & \textbf{84.6} & \textbf{99.8} & \textbf{92.3} \\
\bottomrule
\end{tabularx}
\end{minipage}\hfill
\begin{minipage}[t]{0.405\linewidth}
\centering
\captionof{table}{Normalized and Absolute Average accuracies (\%) of merged full fine-tuned models.}
\label{tab:all_average_fft}
\vspace{0.4em}
\setlength{\tabcolsep}{0.7pt}
\renewcommand{\arraystretch}{1.058}
\rowcolors{5}{}{lightgray}
\begin{tabularx}{\linewidth}{@{} >{\raggedright\arraybackslash\hsize=1.8\hsize}X *{4}{>{\centering\arraybackslash\hsize=0.8\hsize}X} @{}}
\toprule
& \multicolumn{2}{c}{\textbf{ViT-B/32}} & \multicolumn{2}{c}{\textbf{ViT-L/14}} \\
\cmidrule(lr){2-3}\cmidrule(lr){4-5}
\textbf{Method} & \textbf{Norm} & \textbf{Abs} & \textbf{Norm} & \textbf{Abs} \\
\midrule
Zero-shot & 54.8 & 48.1 & 69.8 & 65.3 \\
Indiv. FT & 100.0 & 90.5 & 100.0 & 94.1 \\
\midrule
TA & 76.7 & 69.4 & 88.9 & 83.9 \\
TIES & 81.1 & 73.4 & 90.8 & 85.6 \\
Cons. TA & 81.6 & 74.1 & 90.6 & 85.5 \\
LiNeS & 81.0 & 73.6 & 91.0 & 85.9 \\
FisherAVG & 69.7 & 61.9 & 81.7 & 76.7 \\
RegMean & 87.5 & 79.8 & 93.7 & 88.6 \\
Loc.-Stitch & 86.3 & 79.9 & 90.4 & 86.5 \\
AdaMerge++ & 87.6$^\dagger$ & 81.1$^\dagger$ & 95.1$^\dagger$ & 91.0$^\dagger$ \\
ProDistill & \underline{92.9}$^\dagger$ & \underline{86.0}$^\dagger$ & 96.1$^\dagger$ & 91.9$^\dagger$ \\
TSV & 92.2 & 83.7 & 96.9 & 91.3 \\
ISO-C & 92.4 & 83.8 & \textbf{98.2} & \textbf{92.4} \\
\midrule
\textbf{\methname} & \textbf{94.8} & \textbf{86.1} & \underline{97.6} & \underline{92.0} \\
\bottomrule
\end{tabularx}
\end{minipage}
\end{table}

%% file: chapters/5_analysis.tex
\subsection{Data efficiency of model merging}
\label{sec:data_dependency}
Model merging methods differ in how they use data: weight-space approaches are data-free, apart from the validation data used to tune their hyperparameters, whereas activation-space methods such as RegMean and CoM estimate activation statistics on a small calibration set. At first glance, this reliance on data is the \textbf{main limitation} of our method. We herein assess how critical this dependency is in practice, by varying how much data is available, where it comes from, and what it is used for.
\input{figures/fig_data_dependency}

\textbf{How much data do we need?} \cref{fig:samples_ablation} varies the number of calibration examples per task. CoM is highly data-efficient: already competitive with a handful of samples, it approaches its maximum at roughly $100$ examples, after which gains saturate. A minimum number of examples is thus required to stabilize the Gram matrices of the closed-form solution.

\textbf{If we have labeled data, why not fine-tune?} To assess how efficiently merging exploits calibration data, we use the same $n$ examples per task, drawn from all tasks, to fine-tune three starting points: the pretrained model (the naive choice), the CoM merge, and the best competing merge (\cref{fig:merge_vs_ft}; details in \cref{app:merge_vs_ft}). Initializing from a merged model is consistently more advantageous than starting from the pretrained one, which suggests that the prior of the specialists is preserved and useful. The choice of the initialization matters most in the low-data regime, whereas with more data it matters less, as the data end up compensating for it. Moreover, fine-tuning brings CoM only marginal gains ($+2.2$ and $+0.1$ points at $n=500$ on ViT-B/32 and Llama-3.2-1B) and, with few examples, strongly hurts it, whereas the best competing merge gains much more ($+19.0$ and $+3.0$), showing that CoM already reaches a good optimum, extracting most of the knowledge that the available data can provide.

\Needspace{16\baselineskip}
\input{figures/fig_synthetic}
\textbf{Can we use generated data for calibration?} We further test whether calibration requires real data from the true task distribution, or whether it can be replaced by surrogate data: we sample synthetic images from a generative model and use them for calibration. For each vision task, we generate $500$ images with Stable Diffusion v1.5~\cite{Rombach_2022_CVPR} using one class-conditioned prompt per image, and use them in place of both the calibration and the validation data; the hyperparameters of the data-free baselines are selected on the same synthetic images. On the LoRA checkpoints (\cref{fig:synthetic}; details in \cref{app:synthetic}), CoM remains the best-performing method on both backbones, ahead of the strongest baseline (Core\textsubscript{\tiny ISO}) by $7.4$ points on ViT-B/32 and $4.1$ points on ViT-L/14. Although the margin is smaller than with real data, this suggests that CoM is fairly robust to the choice of the calibration samples.

\subsection{Impact of MCS and feature correlation on merging}
\label{sec:ablation}
\input{figures/fig_ablation_importance}
We quantify the impact of the two main contributions of CoM (correcting Merging Covariate Shift via recursive updates, and feature-correlation weighting) in the most challenging LoRA setting (\cref{tab:reduced_components_ablation}). On both vision backbones, either addressing Merging Covariate Shift or feature-correlation weighting is sufficient to surpass all competing approaches (\cref{tab:all_average_lora}), and the two components are complementary, as their combination yields the best results. On language, feature-correlation weighting brings no gain. We conjecture that the negligible impact of weighting tasks on language is due to these tasks being more similar and equally complex, with a shared label space, whereas vision tasks are very diverse and heterogeneous, e.g., in their out-of-distribution distance from the pretrained model. A more detailed discussion is in \cref{app:components_analysis} and \cref{app:ablation_text_suppl}, and the full ablation including activation normalization is in \cref{tab:variant_components_ablation} of \cref{app:further_imp_details}.

\textbf{Feature correlation.}\label{sec:feature_correlation} To further motivate correlation-based weighting, we test whether our weighting factor $\omega_i^{\,l}$ correlates with a notion of ``\textit{semantic distance}'' between each fine-tuned model and the base model, which we intuitively quantify as the \textbf{performance gap} between the two, i.e., how much fine-tuning improves the task. \cref{fig:importance} compares this gap with our correlation-based weighting factor, averaged across layers to obtain a single value per task\footnote{Both quantities are min-max scaled and $\ell_1$-normalized for visualization; since this is a positive affine transformation, correlations are identical on raw and normalized scores.}. The two measures exhibit a \textbf{clear correlation} across vision datasets and architectures, corroborating our claims (\cref{sec:importance}). Using the pretrained model instead of the merged one to compute the correlation scores yields closely aligned results (see \cref{app:appendix_task_similarity}). On language, instead, tasks are more homogeneous and the correlation becomes weaker (\cref{app:ablation_text_suppl}), consistently with the absence of gains from FC weighting on Llama-3-8B (\cref{tab:reduced_components_ablation}).

\Needspace{17\baselineskip}
\subsection{Computational costs}
\label{sec:computational_costs}
\input{figures/fig_complexity}
We assess computational efficiency (FLOPs) and memory overhead for ViT-B/32 in \cref{fig:complexity}, distinguishing between \texttt{CoM\_fast} (5 samples per task, on par with state-of-the-art methods whose hyperparameters are tuned on the whole validation set) and \texttt{CoM\_best} (500 samples per task). CoM maintains a resource profile \textbf{comparable to other SOTA methods} while being more efficient than RegMean, which requires the entire dataset for calibration. The primary distinction between these two lies in the update rule, which utilizes merged activations rather than task-specific ones. This shift improves accuracy without increasing the number of forward passes (one per task), as activations are cached and the forward pass is effectively paused and resumed at each layer.
We refer the reader to \cref{app:computation_memory_suppl} for experimental results on additional architectures.

%% file: figures/fig_data_dependency.tex
\begin{figure}[t]
\centering
\begin{minipage}[b]{0.42\linewidth}
\centering
\includegraphics[width=\linewidth]{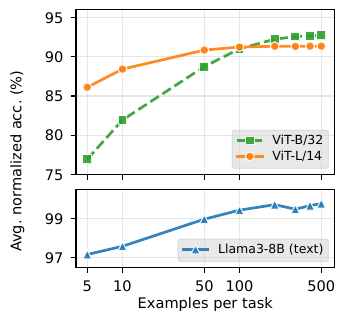}
\captionof{figure}{\textbf{Sample efficiency.} Accuracy of CoM (LoRA) \textit{vs.}\ calibration examples per task, on vision (top) and text (bottom).}
\label{fig:samples_ablation}
\end{minipage}\hfill
\begin{minipage}[b]{0.56\linewidth}
\centering
\includegraphics[width=\linewidth]{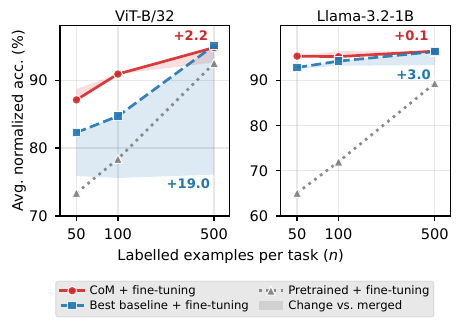}
\captionof{figure}{Fine-tuning after merging on the same $n$ examples per task; shading and numbers: change \textit{vs.}\ the merged model (best baseline picked for each $n$).}
\label{fig:merge_vs_ft}
\end{minipage}
\end{figure}

%% file: figures/fig_synthetic.tex
\setlength{\intextsep}{0pt}

\begin{wrapfigure}[16]{r}{0.44\linewidth}
\centering

\includegraphics[width=\linewidth]{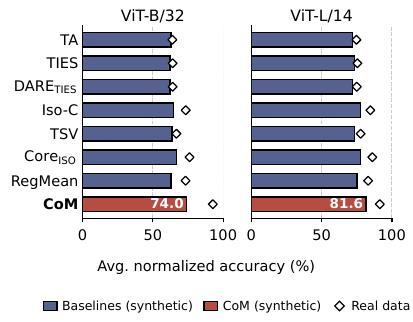}
\caption{Baselines and CoM (LoRA) calibrated on $500$ \textbf{synthetic images} per task (bars); diamonds: methods with real data.}
\label{fig:synthetic}
\end{wrapfigure}

%% file: figures/fig_ablation_importance.tex
\begin{figure}[t]
\centering
\begin{minipage}[t]{0.5\linewidth}
\vspace{0pt}
\centering
\small
\captionof{table}{Impact of CoM components on LoRA models (avg.\ normalized accuracy): \textit{MCS} (covariate shift correction) and \textit{FC} (feature correlation weighting). All rows use activation normalization.}
\label{tab:reduced_components_ablation}
\vspace{0.4em}
\input{tables/reduced_components_ablation}
\end{minipage}\hfill
\begin{minipage}[t]{0.47\linewidth}
\vspace{0pt}
\centering
\includegraphics[width=\linewidth,trim=0 162bp 0 8bp,clip]{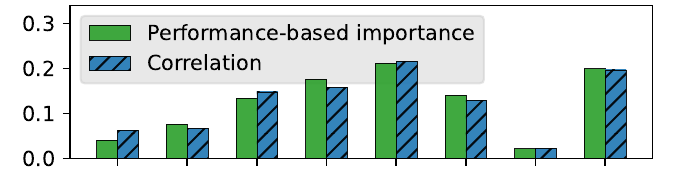}\par\vspace{-1pt}
\includegraphics[width=\linewidth,trim=0 11bp 0 128bp,clip]{images/importance.pdf}
\captionof{figure}{Task importance \textit{vs.}\ feature correlation for ViT-B (above) and ViT-L (below).}
\label{fig:importance}
\end{minipage}
\end{figure}

%% file: tables/reduced_components_ablation.tex
\begingroup
\setlength{\tabcolsep}{3.0pt}
\renewcommand{\arraystretch}{1.0}
\setlength{\extrarowheight}{1pt}
\rowcolors{5}{}{lightgray}

\newcolumntype{Y}{>{\centering\arraybackslash}X}

\begin{tabularx}{\linewidth}{@{}ccYYY@{}}
  \toprule
  \multicolumn{2}{c}{Components}
  & \multicolumn{3}{c}{Architecture} \\
  \cmidrule(lr){1-2} \cmidrule(lr){3-5}
  MCS & FC
  & ViT\mbox{-}B/32
  & ViT\mbox{-}L/14
  & Llama\mbox{-}3\mbox{-}8B \\
  \midrule
  \xmark & \xmark & 74.32 & 83.05 & 97.52 \\
  \cmark & \xmark & 85.83 & 87.38 & 99.76 \\
  \xmark & \cmark & 84.02 & 88.27 & 97.52 \\
  \cmark & \cmark & 92.74 & 91.35 & 99.63 \\
  \bottomrule
\end{tabularx}

\endgroup

%% file: figures/fig_complexity.tex
\begin{wrapfigure}[17]{r}{0.48\linewidth}
    \centering
    \vspace{-0.2em}
    \includegraphics[width=\linewidth]{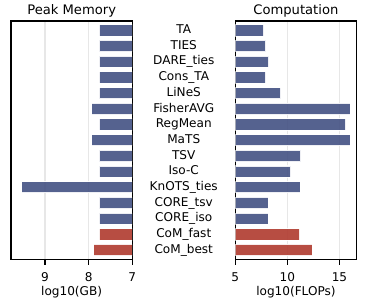}
    \vspace{-0.8em}
    \caption{Peak memory (left), FLOPs (right) on ViT-B/32.}
    \label{fig:complexity}
    \vspace{0.6em}
\end{wrapfigure}

%% file: chapters/6_related.tex
\section{Comparison with Related Work}
\label{sec:related}
Early model merging methods average checkpoints, as in FedAvg~\cite{mcmahan2017communication}, Model Soups~\cite{wortsman2022model}, and WiSE-FT~\cite{wortsman2022robust}, or add task vectors to the pretrained weights, as in Task Arithmetic~\cite{ilharco2023editing}, showing that independently fine-tuned models can often be combined without retraining. However, heterogeneous task updates interfere with each other~\cite{tang2024merging,tam2023merging}. TIES~\cite{yadav2023ties} mitigates this by trimming small updates and resolving sign conflicts, whereas DARE~\cite{yu2024language} randomly drops and rescales weights. Consensus~\cite{wang2024localizing} keeps only weights relevant to multiple tasks, and LiNeS~\cite{wang2025lines} scales the updates with layer depth.

A second family merges in aligned parameter subspaces: TSV~\cite{gargiulo2025task}, KnOTS~\cite{stoica2025model}, Core\textsubscript{\tiny TSV} and Core\textsubscript{\tiny ISO}~\cite{panariello2025accurate}, and Iso-C~\cite{marczak2025no} use SVD, low-rank compression, whitening, or isotropic rescaling to align task updates. These methods are effective and often efficient, but their alignment remains defined in parameter space. CoM is complementary: it matches layer behavior directly on activations, updating them as the merged network is built.

A complementary line of merging-aware training instead biases training toward solutions that are mergeable by design. Tangent-space fine-tuning~\cite{ortizjimenez2023tangent} linearizes the model around the pretrained weights, so that a linear combination of parameters corresponds to a linear combination of model outputs; $\tau$Jp~\cite{yoshida2025mastering} promotes weight disentanglement by penalizing representation drift, and TAK~\cite{porrello2026dataless} does so without access to other tasks' data, through a dataless regularizer built on a Kronecker-factorized approximation of the Generalized Gauss-Newton matrix. Unlike these approaches, CoM leaves training untouched and operates post hoc on independently fine-tuned models.

Feature- and activation-based methods are closest to our work: Neuron Alignment~\cite{tatro2020optimizing}, ZipIt!~\cite{stoica2023zipit}, Optimal Transport Fusion~\cite{singh2020model}, Fisher-weighted averaging~\cite{matena2022merging}, RegMean~\cite{jin2023dataless}, LoRM~\cite{salami2025closed}, MaTS~\cite{tam2023merging}, ProDistill~\cite{xu2025scalable}, and AdaMerging~\cite{yang2023adamerging} use importance, activation matching, or data-driven optimization to improve compatibility. However, they typically estimate layer-wise statistics from the original task-specific networks, or optimize layer coefficients, without correcting the distribution shift induced by previously merged layers. CoM instead recomputes activation statistics sequentially, making layer-wise matching conditionally consistent.

Another line of work aligns and recalibrates networks before combining them. Permutation-based re-basin methods~\cite{ainsworth2023git,pena2023rebasin}, including TransFusion~\cite{rinaldi2025update} for task vectors, align the units of independently trained models into the same loss basin, and REPAIR~\cite{jordan2022repair} corrects the internal activation statistics of the interpolated network, which collapse when two models' weights are averaged. These methods couple adjacent layers, as the alignment of a layer is constrained by its neighbors, but typically align one model at a time to a single target. The dependency modeled by CoM is instead inherently multi-task: once an earlier layer is merged, the inputs of the following layers change differently for each task, so statistics collected from the original task-specific networks become stale, and each merged layer must accommodate several shifted input distributions at once.

Finally, dynamic merging approaches adapt the merged weights to the task or input at inference time. EMR-Merging~\cite{huang2024emr} applies lightweight task-specific masks and rescaling factors to a unified task vector, given the task identity; Twin-Merging~\cite{lu2024twin} and WeMoE~\cite{tang2024merging} instead use a router to combine task-specific modules or task vectors for each input. These methods gain accuracy at the price of additional storage, task information, or routing at inference. We instead follow the standard static inference protocol: a single merged backbone serves all tasks, without task identifiers or routing.

%% file: chapters/7_conclusions.tex
\section{Discussion}
In this work, we identify and address Merging Covariate Shift (MCS), a form of internal covariate shift that emerges when merging layers independently in methodologies that rely on activation statistics. To mitigate the adversarial effect of this phenomenon, we propose Chain of Merges (\methname), a framework that updates activation statistics autoregressively, capturing inter-layer dependencies and fully eliminating MCS. Empirical results on standard vision and language benchmarks demonstrate that \methname consistently outperforms existing methods across diverse architectures and domains.

\textbf{Limitations.}\label{sec:limitations} CoM requires calibration examples (albeit only a few) to estimate activation statistics, and its gains are less pronounced with full fine-tuning on vision tasks, where merging is inherently easier and benchmark performance is nearing saturation.

%% file: supplementary/0_supplementary.tex
\appendix
\section*{\LARGE Appendix}

The appendix is organized as follows:
\begin{itemize}
    \item \cref{app:group_method} collects the details of the method:
    \begin{itemize}
        \item \cref{app:math_suppl} provides the closed-form derivations: the correlation-weighted merging solution (\cref{app:derivation_weighted}) and the merging of normalization layers (\cref{app:derivation_layernorm}).
        \item \cref{app:further_imp_details} gives further implementation details: the activation normalization that improves the conditioning of the Gram matrices, and the complete algorithm.
        \item \cref{app:computation_memory_suppl} compares the computational (FLOPs) and memory cost of CoM against existing methods.
    \end{itemize}
    \item \cref{app:group_analysis} collects further analyses of CoM:
    \begin{itemize}
        \item \cref{app:components_analysis} complements the ablation of the main paper with a further analysis of the components of CoM (MCS correction, feature-correlation weighting, and activation normalization).
        \item \cref{app:appendix_task_similarity} analyzes the task importance coefficients, covering their theoretical motivation (\cref{app:orthogonality_generalization}) and an ablation on the reference model used to compute them (\cref{app:reference_models}).
        \item \cref{app:ablation_text_suppl} reports supplementary ablations on the textual datasets, namely the correlation-based importance and the MCS analysis on Llama-3-8B.
        \item \cref{app:regime_discussion} discusses why the gains of CoM are larger with LoRA than with full fine-tuning.
    \end{itemize}
    \item \cref{app:group_data} collects the experiments on the dependency on data:
    \begin{itemize}
        \item \cref{app:calibration_sensitivity} studies the sensitivity of CoM to the construction of the calibration set (sampling strategy, data split, input noise, class coverage, and the numerical values of the sample-efficiency plot).
        \item \cref{app:synthetic} evaluates calibration on synthetic images generated with Stable Diffusion v1.5, when no real task data is available.
        \item \cref{app:merge_vs_ft} compares merging with fine-tuning when the same examples are available.
    \end{itemize}
    \item \cref{app:group_results} collects additional results:
    \begin{itemize}
        \item \cref{app:more_accuracies} lists the accuracy on each dataset for ViT-B/32, ViT-L/14, and Llama-3-8B with LoRA fine-tuning.
        \item \cref{app:fft_language} reports full fine-tuning results on the language tasks.
    \end{itemize}
\end{itemize}

\clearpage
\section{Method Details}
\label{app:group_method}
\subsection{Closed-form derivations}
\label{app:math_suppl}
\subsubsection{Derivation of the correlation-weighted merging solution}
\label{app:derivation_weighted}
We want to minimize the following objective with respect to the matrix
$\mW_M^{\,l}$:
\begin{equation*}
\widehat{\Omega}_l=\sum_{i=1}^N \frac{\omega_i^{\,l}}{T_i}\,
\|\mW_M^{\,l}\tilde{\mX}_i^{\,l}-\mW_i^{\,l}\tilde{\mX}_i^{\,l}\|_{F}^2.
\end{equation*}
To simplify the notation, define $\mG_i^{\,l}=\frac{1}{T_i}\tilde{\mX}_i^{\,l}(\tilde{\mX}_i^{\,l})^\top$.
Expanding the norm and applying standard rules of matrix differentiation, the gradient of the objective with respect to $\mW_M^{\,l}$ is:
\begin{equation*}
\nabla_{\mW_M^{\,l}} \widehat{\Omega}_l
=2\Biggl[
   \mW_M^{\,l}\Bigl(\sum_{i=1}^N \omega_i^{\,l}\mG_i^{\,l}\Bigr)
   -\Bigl(\sum_{i=1}^N \omega_i^{\,l}\mW_i^{\,l}\mG_i^{\,l}\Bigr)
 \Biggr].
\end{equation*}
The minimizer is obtained by setting the derivative equal to zero as:
\begin{equation*}
\mW_M^{\,l}\Bigl(\sum_{i=1}^N \omega_i^{\,l}\mG_i^{\,l}\Bigr)
=\sum_{i=1}^N \omega_i^{\,l}\mW_i^{\,l}\mG_i^{\,l}.
\end{equation*}
Finally, we can solve for $\mW_M^{\,l}$ by multiplying on the right hand side with the inverse of $\sum_{i=1}^N \omega_i^{\,l}\mG_i^{\,l}$:
\begin{equation*}
\mW_M^{\,l}
=\Biggl(\sum_{i=1}^N \omega_i^{\,l}\mW_i^{\,l}\mG_i^{\,l}\Biggr)
 \Biggl(\sum_{i=1}^N \omega_i^{\,l}\mG_i^{\,l}\Biggr)^{-1}.
\end{equation*}
If the latter matrix is singular, the Moore--Penrose pseudoinverse should be used instead.

\subsubsection{Closed-form merging of normalization layers}
\label{app:derivation_layernorm}

Transformer blocks contain element-wise normalization layers (LayerNorm, RMSNorm) parameterized by a per-channel multiplicative scale $\gamma$ and additive bias $\beta$. Since each output channel is processed independently, the activation-matching objective decouples into a collection of scalar problems -- one per channel and per layer -- and admits a simple closed-form solution: the merged parameters reduce to the arithmetic mean of the task-specific ones, independently of the underlying activation distribution. Derivation for the multiplicative scale below; the additive case is analogous.

\textbf{Setup.} Fix a single channel of a normalization layer and let $x \in \mathbb{R}$ denote its input scalar activation, after the standard normalization performed by the layer itself. Each task-specific model $i \in \{1, \dots, N\}$ applies a learned scale $\gamma_i$, producing $\gamma_i\, x$. We seek the merged scale $\gamma_m$ that minimizes the expected squared deviation from the average output of the individual models on this channel.

\textbf{Multiplicative scale.} The matching objective is
\[
\arg\min_{\gamma_m}\; \mathbb{E}_x\!\left[\left(\gamma_m\, x - \frac{1}{N}\sum_{i=1}^{N} \gamma_i\, x\right)^2\right].
\]
Factoring out $x$ and pulling the data-independent term outside the expectation gives
\[
\arg\min_{\gamma_m}\; \left(\gamma_m - \frac{1}{N}\sum_{i=1}^{N}\gamma_i\right)^{\!2} \mathbb{E}_x[x^2].
\]
Assuming the channel is not trivially dead, $\mathbb{E}_x[x^2] > 0$, the global minimum is attained at
\[
\gamma_m \;=\; \frac{1}{N}\sum_{i=1}^{N}\gamma_i,
\]
independently of the underlying activation distribution.

\textbf{Additive bias.} An analogous derivation (in which $x$ does not appear because the term enters additively) yields
\[
\beta_m \;=\; \frac{1}{N}\sum_{i=1}^{N}\beta_i.
\]

\textbf{Implication.} Merging normalization layers therefore requires neither activation statistics nor an iterative procedure: the optimal merged $\gamma$ and $\beta$ are simply the per-channel arithmetic means of the task-specific parameters. This formally justifies our use of plain averaging for all non-linear layers reported in the implementation details, without sacrificing optimality with respect to the activation-matching objective.

\subsection{Further Implementation Details}
\label{app:further_imp_details}
\input{tables/variant_components_ablation}
\input{algorithms/chain_of_merges}
\paragraph{Activation Normalization.} Our proposed approach requires the inversion of $\sum_{i=1}^N \widehat{\mX}_i^l \widehat{\mX}_i^{l\Ttop}$, whose conditioning critically affects the numerical stability of the solution. In Transformer architectures, activation Gram matrices are frequently ill-conditioned due to the inherent low-dimensionality of the underlying token representations~\citep{barbero2406transformers,arefin2024seq}, especially as fine-tuning typically occurs within a low-dimensional subspace~\citep{aghajanyan-etal-2021-intrinsic,kumar2022fine}. Drawing on prior work showing that layer normalization~\citep{ba2016layer} mitigates representational collapse~\citep{wu2024role}, we replace the Gram matrix $\widehat{\mX}_i^l \widehat{\mX}_i^{l\Ttop}$ with the correlation matrix $\mG^{\,l}_i$, which is computed from the per-token normalized features $\tilde{\mX}_i^l$, averaged over the $T_i$ tokens, and therefore better conditioned. The full ablation study in terms of average normalized accuracy comprising the effect of the activation normalization is in \cref{tab:variant_components_ablation}: normalizing the activations always benefits performance, especially on ViT-L/14, with an increase of over $3$ points. The complete procedure of the proposed methodology is summarized in \cref{alg:chain_merges}.

\subsection{Computational and Memory Cost}
\label{app:computation_memory_suppl}
\begin{figure}[H]
    \centering

    \begin{subfigure}{0.9\textwidth}
        \centering
        \includegraphics[width=0.8\linewidth]{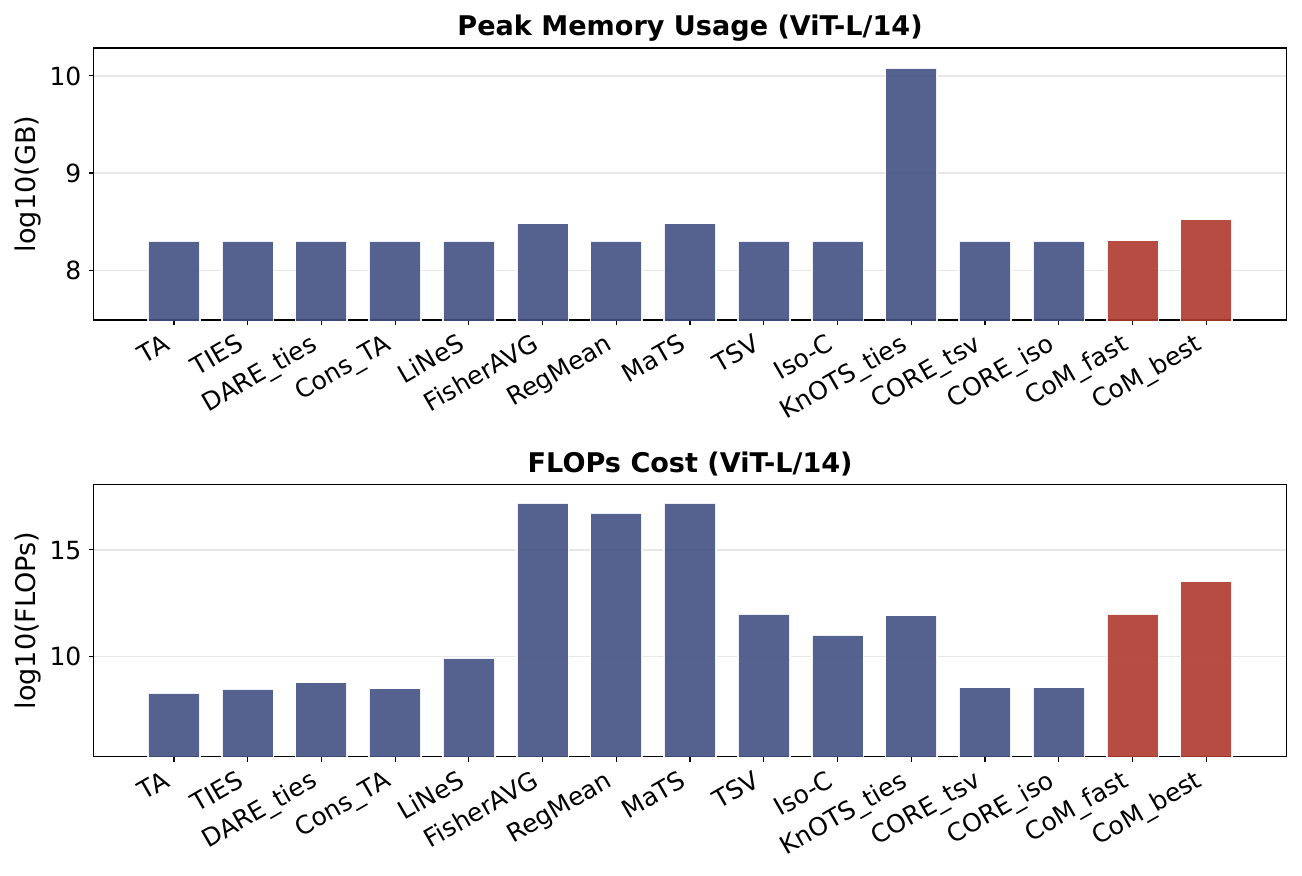}

        \caption{Peak memory usage and merging computational cost for ViT-L/14.}
        \label{fig:vitl}
    \end{subfigure}

    \vspace{1em}
    \begin{subfigure}{0.9\textwidth}
        \centering
        \includegraphics[width=0.8\linewidth]{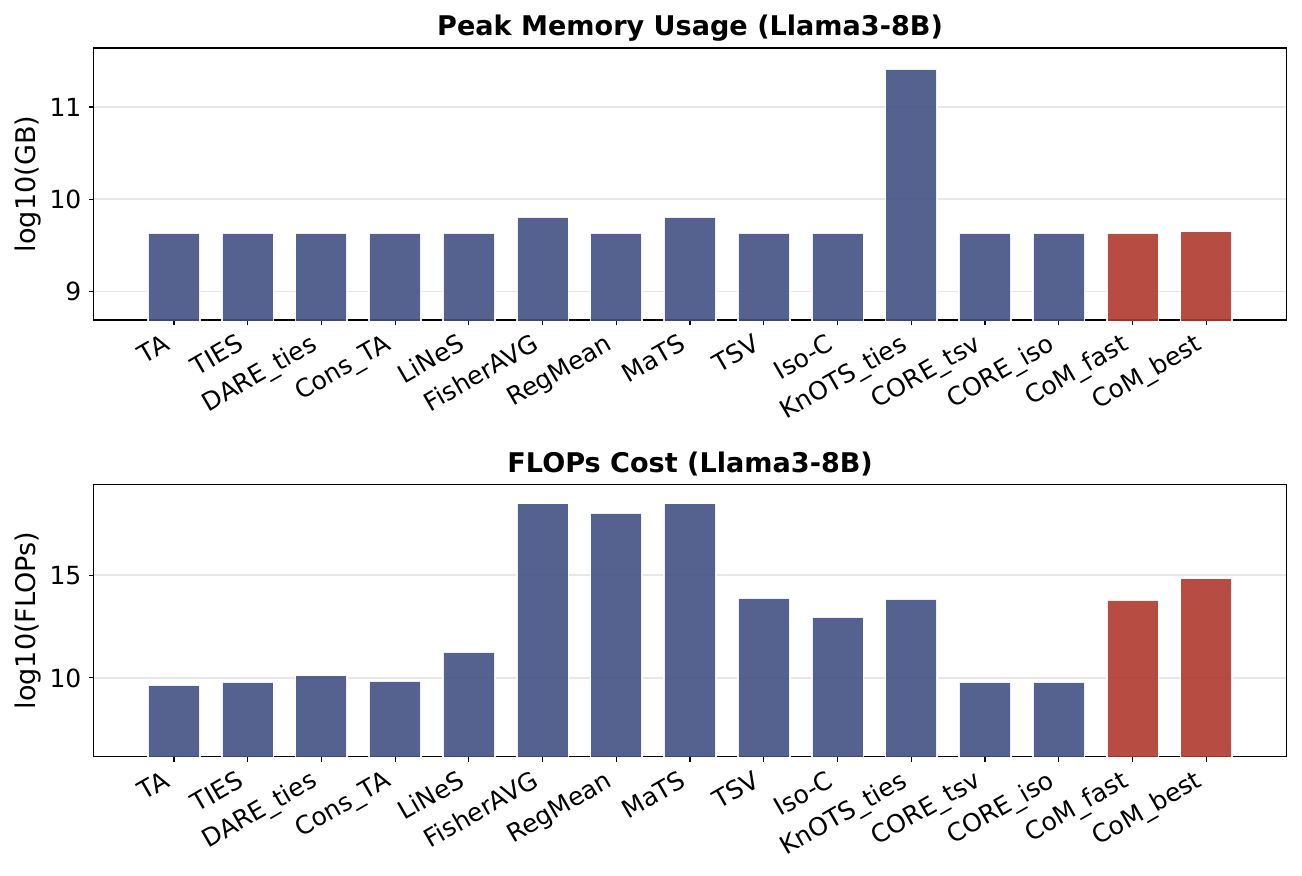}

        \caption{Peak memory usage and merging computational cost for Llama-3-8B.}
        \label{fig:llama}
    \end{subfigure}

    \caption{Comparison of merging methods across computational and memory cost.}
\end{figure}

We evaluate the efficiency of the proposed methodology in terms of computational complexity and memory usage, comparing it against established techniques. For the architectures used\footnote{We refer the reader to \cref{fig:complexity} for ViT-B/32.} \cref{fig:vitl,fig:llama} (log-scale) report the number of FLOPs as a measure of computational cost, and the theoretical memory overhead in GB. We denote our methodology with a minimal calibration set of $5$ samples per task, with which CoM is already on par with state-of-the-art methods whose hyperparameters are tuned on the whole validation set, as \texttt{CoM_fast}, and the variant using the optimal number of samples for maximum performance as \texttt{CoM_best}. The results show that CoM has a computational and memory cost comparable to alternative merging strategies.

A notable comparison is with RegMean. Although our approach shares a similar formulation, it differs in the update rule by using the merged activations instead of the task-specific ones. This key distinction improves accuracy while keeping the number of forward passes unchanged: RegMean requires one forward pass per task-specific model to compute the Gram matrices, whereas CoM performs the same number of passes on the merged model. Finally, CoM requires fewer examples in practice, resulting in consistently lower computational cost, as shown in \cref{fig:samples_ablation}. These findings highlight that the proposed recursive scheme provides an effective balance between efficiency and performance, ensuring state-of-the-art accuracy while keeping both computation and memory overhead limited.

\section{Further Analysis of CoM}
\label{app:group_analysis}
\subsection{Further Analysis of the Components of CoM}
\label{app:components_analysis}
This section complements the ablation study of \cref{sec:ablation}. In \cref{tab:reduced_components_ablation,tab:variant_components_ablation}, the rows with MCS$\,=\,$\xmark\ are layer-wise solutions computed on the activations of each task's own fine-tuned network, using the same class-balanced calibration samples as CoM; all values are averaged over three calibration seeds. Correcting MCS yields consistent gains over the layer-wise solution without chain on all three architectures, even without activation normalization ($+10.7$, $+3.6$ and $+2.4$ points on ViT-B/32, ViT-L/14 and Llama-3-8B; \cref{tab:variant_components_ablation}), whereas FC weighting is mostly beneficial on vision and can even hurt on language when feature normalization is off. Equipping the layer-wise solution with the activation normalization used by CoM brings comparatively small gains on all three architectures (\cref{tab:variant_components_ablation}), indicating that the improvement stems from the recursion rather than from normalization. The components are also complementary on vision: on ViT-L/14, for instance, the full model improves over MCS alone by about $4$ points and over FC alone by about $3$ points.

\subsection{Analysis of Task Importance Coefficients}
\label{app:appendix_task_similarity}

In this section, we provide a deeper analysis of CoM's task importance coefficients. We first elaborate on the theoretical connection between feature correlation and generalization, and subsequently validate our choice of using the merged model as a reference for computing these statistics.

\subsubsection{Theoretical Motivation: Orthogonality and Generalization}
\label{app:orthogonality_generalization}
Our approach quantifies inter-feature correlation to estimate task importance. We argue that, for in-distribution inputs, representations from large-scale pretrained models are approximately decorrelated, as they capture general-purpose structures rather than task-specific patterns.

Consequently, high off-diagonal correlations in the input Gram matrix $\mG$ after fine-tuning indicate that the task concentrates on a narrow subspace of the original data distribution, diverging significantly from the pretraining initialization. Intuitively, when a task heavily relies on a limited set of feature directions, those features exhibit higher correlation, revealing that the model is focusing on a restricted region of the representation space. This perspective aligns with established literature demonstrating that feature decorrelation is linked to improved generalization \citep{cogswell2015reducing, morcos2018importance}. Our method leverages this principle: we treat significant deviations from the decorrelated pretrained state (high inter-feature correlation) as a signal of task specificity.

\subsubsection{Ablation on Reference Models}
\label{app:reference_models}
\begin{figure}[t]
    \centering
    \includegraphics[width=\linewidth]{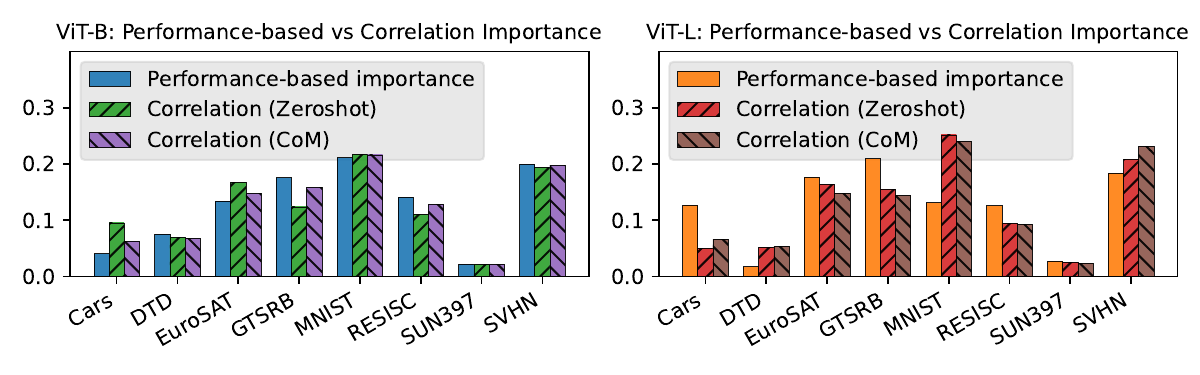}
    \caption{\textbf{Ablation of Task Importance Coefficients (ViT-B and ViT-L).} We compare the importance scores derived from our proposed CoM method (which uses the merged model as a reference) against those derived from the zero-shot pretrained model and a performance-based oracle. The results demonstrate that CoM coefficients strongly correlate with the zero-shot reference, justifying the use of the merged model as an efficient proxy.}
    \label{fig:importance_ablation}
\end{figure}
To maintain comparability across tasks and layers, correlation coefficients must be computed using a single, task-agnostic reference model. Using task-specific fine-tuned models as references would be improper, as this would quantify how well a model fits its own specific task rather than providing a standardized measure of distributional shift.

Ideally, the zero-shot (pretrained) model serves as the ground truth reference. However, for computational efficiency within our pipeline, CoM computes correlations using the merged model. We posit that the merged model is a valid proxy because it remains approximately task-agnostic and provides a balanced representation across all tasks. To validate this approximation, we conducted an ablation study comparing the task importance scores derived from three different sources:
\begin{enumerate}
    \item \textbf{Performance-based importance (Oracle):} importance scores calculated based on actual evaluation performance.
    \item \textbf{Correlation (Zero-shot):} coefficients computed using the original pretrained checkpoint (the ideal reference).
    \item \textbf{Correlation (CoM):} coefficients computed using our proposed method with the merged model (the efficient proxy).
\end{enumerate}

The results, illustrated in \cref{fig:importance_ablation} for both ViT-B and ViT-L architectures, show a high degree of alignment between the coefficients computed via CoM (purple/brown hatched bars) and those computed via the zero-shot model (green/red hatched bars). This confirms that using the merged model to compute the Gram matrices does not introduce significant deviation from the ideal pretrained reference, validating the efficiency of our protocol. Furthermore, both correlation-based metrics track the general trends of the performance-based oracle, particularly in identifying high-importance tasks such as EuroSAT and SVHN.

\subsection{Supplementary Ablations for Textual Datasets}
\label{app:ablation_text_suppl}

\begin{figure}[H]
    \centering
    \begin{minipage}[b]{0.43\textwidth}
        \centering
        \includegraphics[width=\linewidth]{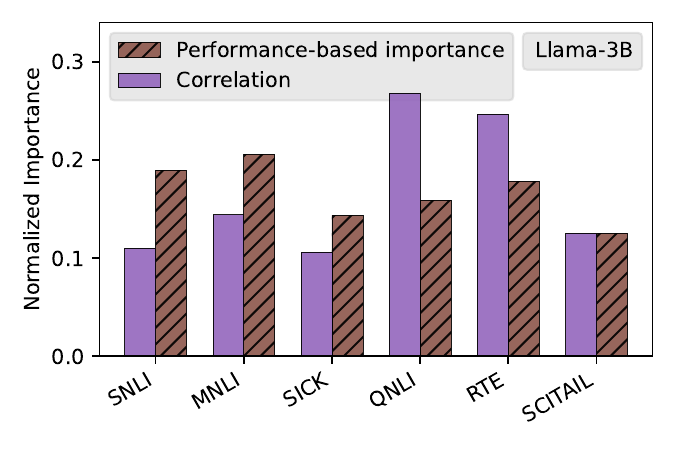}
        \caption{Task importance \textit{vs.}\ correlation.}
        \label{fig:importance_suppl}
    \end{minipage}
    \hspace{2em}
    \vspace{0.5em}
    \begin{minipage}[b]{0.41\textwidth}
        \centering{\includegraphics[width=\linewidth]{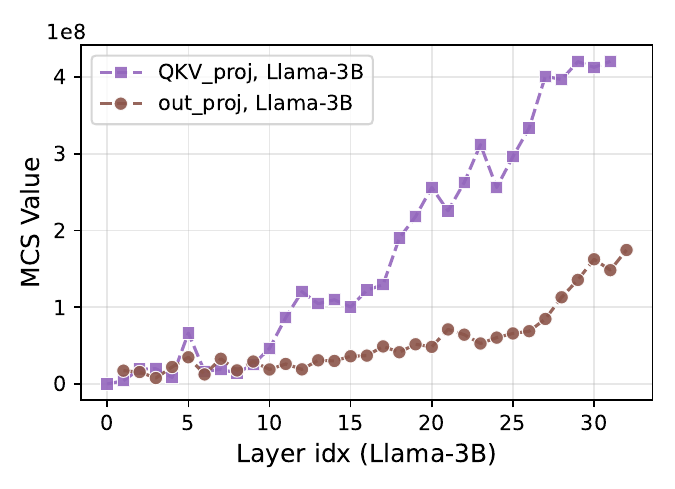}}
        \caption{Merging Covariate Shift across layers.}
        \vspace{-0.5em}
        \label{fig:mcs_suppl}
    \end{minipage}
\end{figure}

\paragraph{Correlation-based importance.} In \cref{fig:importance_suppl}, we report the correlation-based importance analysis for textual tasks (SNLI, MNLI, SICK, QNLI, RTE, and SciTail). Following the approach used for vision datasets, we compute the performance gap between task-specific fine-tuned models and the zero-shot performance of the base Llama-3-8B model, and compare it to the correlation-based weighting factor $\omega_i^{\,l}$ averaged across layers. In contrast to vision tasks, textual datasets exhibit a relatively uniform importance distribution: both performance gaps and correlation weights vary little across tasks. As a result, the correlation between the two measures is weaker, and correlation-weighted merging provides limited benefit. These findings highlight that the effectiveness of correlation-based weighting depends on task heterogeneity, as greater diversity amplifies its impact on the merged model.

\paragraph{Measuring MCS.} In \cref{fig:mcs_suppl}, we extend the analysis from \cref{sec:ablation} to the Llama-3-8B architecture, computing the MCS for each layer using the same methodology. It can be seen that the amount of Merging Covariate Shift observed in textual datasets is broadly comparable to that identified in vision models. This consistency suggests that the underlying dynamics of parameter interference and shift remain stable across different model scales and modalities.

\subsection{Larger Gains with LoRA}
\label{app:regime_discussion}
Both our results and prior work~\citep{stoica2025model,zeng2025robustmerge} suggest that LoRA checkpoints are intrinsically harder to merge than fully fine-tuned ones. In the full-rank weight space, task-specific solutions can distribute their updates across broader and more aligned subspaces, which reduces interference during merging and compresses the gap between methods. In contrast, LoRA constrains adaptation to a small number of directions, and parameter-efficient updates suffer from direction instability among singular values~\citep{zeng2025robustmerge}: different tasks are therefore more likely to encode their knowledge within overlapping dominant directions of the pretrained model, which increases competition and task interference. When interference is more severe, as with LoRA, errors introduced in early layers propagate through a restricted adaptation space, and correcting MCS becomes particularly important. This explains why the largest gains of CoM emerge in the LoRA regime. In addition, on ViT-L/14 with full fine-tuning the benchmark is essentially saturated (Iso-C: $98.2\%$ \textit{vs.}\ CoM: $97.6\%$ normalized accuracy), leaving little room for significant gains.

\section{Data Dependency}
\label{app:group_data}
\subsection{Sensitivity to the Calibration Set}
\label{app:calibration_sensitivity}
We analyze the dependency of CoM on the construction of the calibration set (500 samples per task; by default, from the training split on Llama-3-8B, whose RTE and SICK validation splits are smaller) by comparing: (i) class-balanced \textit{vs.}\ unconstrained random sampling, (ii) samples drawn from the validation \textit{vs.}\ the training split, (iii) calibration inputs perturbed with Gaussian noise of standard deviation $\sigma_{\text{noise}}$ added to the preprocessed images, and (iv) calibration restricted to at most $3\%$ of the classes of each task (a single class for the 10-class datasets). \cref{tab:calibration_sensitivity} reports the average normalized accuracy over three seeds. Noise is added to the input embeddings with standard deviation $\sigma_{\text{noise}}$ times their root-mean-square.

\begin{table}[ht]
\centering
\caption{Average normalized accuracy (\%) of CoM (LoRA) under different calibration set constructions. Mean $\pm$ standard deviation over 3 seeds.}
\label{tab:calibration_sensitivity}
\rowcolors{2}{}{lightgray}
\setlength{\tabcolsep}{0.55em}
\begin{tabular}{lccc}
\toprule
Calibration set & ViT-B/32 & ViT-L/14 & Llama-3-8B \\
\midrule
Class-balanced, validation split (default, vision) & 92.74 $\pm$ 0.14 & 91.35 $\pm$ 0.02 & 99.57 $\pm$ 0.15 \\
Random, validation split & 92.60 $\pm$ 0.06 & 91.31 $\pm$ 0.03 & 99.88 $\pm$ 0.12 \\
Class-balanced, training split (default, Llama) & 92.53 $\pm$ 0.05 & 91.30 $\pm$ 0.03 & 99.76 $\pm$ 0.12 \\
Random, training split & 92.39 $\pm$ 0.06 & 91.25 $\pm$ 0.01 & 99.69 $\pm$ 0.10 \\
Noise $\sigma_{\text{noise}}=0.01$ & 90.64 $\pm$ 0.13 & 85.52 $\pm$ 0.14 & 99.72 $\pm$ 0.11 \\
Noise $\sigma_{\text{noise}}=0.1$ & 82.05 $\pm$ 0.13 & 81.61 $\pm$ 0.23 & 99.70 $\pm$ 0.12 \\
Noise $\sigma_{\text{noise}}=1.0$ & 69.10 $\pm$ 0.63 & 67.44 $\pm$ 2.25 & 99.19 $\pm$ 0.07 \\
At most $3\%$ of the classes & 79.23 $\pm$ 0.58 & 84.75 $\pm$ 0.61 & 98.36 $\pm$ 0.42 \\
\bottomrule
\end{tabular}
\end{table}

\vspace{1em}

On all architectures, the sampling strategy has a limited impact: using the training instead of the validation split, or random instead of class-balanced sampling, changes the accuracy by at most $0.4$ points on ViT-B/32, $0.1$ points on ViT-L/14 and $0.2$ points on Llama-3-8B. Performance degrades progressively as the calibration inputs are corrupted with noise, and the largest drops occur under severe distributional mismatch, i.e., strong noise or calibration on very few classes. On Llama-3-8B, noise perturbation does not substantially impact performance.

\paragraph{Number of calibration examples.} \cref{tab:samples_ablation_full} reports the numerical values of \cref{fig:samples_ablation} for all the sample sizes of CoM's calibration set, from $5$ to $500$ examples per task.
\input{tables/samples_ablation_full}

\subsection{Synthetic Calibration Data}
\label{app:synthetic}
In this section, we evaluate merging methods in a setting where no real task data is available, but samples can be generated. For each vision task, we generate $500$ images with Stable Diffusion v1.5~\cite{Rombach_2022_CVPR}, one class-conditioned prompt per image, at $512\times512$ and downsampled to the native resolution of the dataset. The generated images replace the real data: RegMean and CoM are calibrated on all synthetic images and use their default hyperparameters, while the hyperparameters of the data-free methods are selected on the same synthetic images instead of the real validation split. \cref{tab:synthetic} reports the results on the LoRA checkpoints (single run). CoM remains the best method on both backbones, ahead of the strongest baseline (Core\textsubscript{\tiny ISO}) by $7.4$ points on ViT-B/32 and $4.1$ points on ViT-L/14, although the margin is smaller than with real data (\cref{tab:all_average_lora}).

\begin{table}[ht]
\centering
\caption{Average normalized accuracy (\%) of merged LoRA models using synthetic data ($500$ generated images per task) in place of the real calibration and validation data.}
\label{tab:synthetic}
\begingroup
\setlength{\tabcolsep}{4pt}
\rowcolors{2}{}{lightgray}
\setlength{\tabcolsep}{0.75em}
\begin{tabular}{lcccccccc}
\toprule
Arch. & TA & TIES & DARE\textsubscript{\tiny TIES} & Iso-C & TSV & Core\textsubscript{\tiny ISO} & RegMean & CoM \\
\midrule
ViT-B/32 & 62.91 & 62.27 & 62.21 & 64.73 & 63.63 & \underline{66.64} & 62.95 & \textbf{74.01} \\
ViT-L/14 & 72.14 & 73.25 & 71.97 & 77.52 & 73.18 & \underline{77.53} & 75.24 & \textbf{81.59} \\
\bottomrule
\end{tabular}
\endgroup
\end{table}
\vspace{1em}

\subsection{Merging \textit{vs.} Fine-Tuning with the Same Data}
\label{app:merge_vs_ft}
Since CoM needs a few examples per task for merging, one may ask whether the same examples would be better spent on tuning the hyperparameters of a data-free method or on fine-tuning a merged model. We address this question with a matched-data protocol on the LoRA checkpoints of ViT-B/32 and ViT-L/14 and, since repeatedly fine-tuning Llama-3-8B is too expensive, on Llama-3.2-1B fine-tuned on the six language tasks with LoRA adapters (rank $16$ on the attention projections). For $n \in \{10, 50, 100, 500\}$, we draw $n$ examples per task uniformly at random from the split used for calibration in the rest of the paper (validation for vision, training for language; not class-balanced, hence the slight differences in the CoM results w.r.t.\ \cref{tab:all_average_lora,tab:samples_ablation_full}), and every method receives exactly the same examples.

CoM and RegMean are calibrated on all $n$ examples with their default hyperparameters and use no labels; RegMean is therefore run on $n$ examples here, rather than on the entire training set as in its original setting. The hyperparameters of all other methods are tuned on the same $n$ labeled examples (indicated as \textit{Merged} in \cref{tab:merge_vs_ft}).

We then fine-tune the backbone of every merged model on the same examples (\textit{+\,Fine-tuning}): $80\%$ of the examples are used for the gradient updates and the remaining $20\%$ for early stopping (patience $5$, at most $1000$ steps) and for selecting the learning rate in $\{10^{-7}, \dots, 10^{-4}\}$. As a reference without merging, the same fine-tuning is also applied to the pretrained backbone (\textit{Zero-shot}). Not fine-tuning is also a candidate: the fine-tuned model is retained only if it improves over the merged model on the early-stopping examples, and the merged model is reported otherwise. We use AdamW, keep the classification heads frozen (on Llama-3.2-1B, all transformer blocks are fine-tuned and the embeddings are frozen as well), and evaluate the test split once, for the selected model. Results are averaged over three seeds that fix the sampled examples.

\begin{table}[ht]
\centering
\caption{Average normalized accuracy (\%) of LoRA merged models when all methods use the same $n$ examples per task (mean over 3 seeds).}
\label{tab:merge_vs_ft}
\begingroup
\setlength{\tabcolsep}{0.95em}
\rowcolors{5}{}{lightgray}
\begin{tabular}{lcccccccc}
\toprule
& \multicolumn{4}{c}{Merged} & \multicolumn{4}{c}{+\,Fine-tuning} \\
\cmidrule(lr){2-5} \cmidrule(lr){6-9}
Method & $n\!=\!10$ & $50$ & $100$ & $500$ & $n\!=\!10$ & $50$ & $100$ & $500$ \\
\midrule
\rowcolor{white}\multicolumn{9}{l}{\textit{ViT-B/32}} \\
\rowcolor{white}Zero-shot & 60.6 & 60.6 & 60.6 & 60.6 & 63.8 & 73.4 & 78.4 & 92.5 \\
\cmidrule(lr){1-9}
Task Arithmetic & 63.6 & 63.2 & 63.9 & 63.9 & 66.9 & 75.3 & 80.1 & 93.6 \\
TIES & 63.6 & 63.9 & 63.8 & 64.0 & 67.3 & 75.6 & 80.3 & 93.4 \\
DARE-TIES & 64.1 & 63.9 & 64.1 & 64.1 & 67.8 & 75.8 & 80.4 & 93.6 \\
LiNeS & 63.6 & 63.2 & 63.9 & 63.9 & 66.9 & 75.3 & 80.1 & 93.6 \\
Consensus & 64.2 & 65.1 & 65.1 & 65.3 & 65.8 & 77.3 & 81.3 & 92.8 \\
TSV & 66.8 & 66.5 & 66.5 & 66.5 & 72.2 & 77.3 & 82.1 & 93.6 \\
KnOTS-TIES & 67.5 & 68.1 & 68.1 & 68.3 & 72.0 & 78.9 & 82.0 & 94.3 \\
Iso-C & 69.9 & 73.2 & 73.2 & 73.1 & 69.2 & 81.1 & 84.3 & 94.4 \\
Core\textsubscript{\tiny TSV} & 66.6 & 66.6 & 66.6 & 66.6 & 70.9 & 78.2 & 81.7 & 93.7 \\
Core\textsubscript{\tiny ISO} & \underline{76.0} & \underline{75.9} & \underline{75.6} & \underline{76.1} & \underline{76.0} & \underline{82.3} & \underline{84.7} & \textbf{95.1} \\
RegMean & 72.4 & 73.3 & 72.9 & 72.9 & 75.3 & 81.6 & 84.0 & \textbf{95.1} \\
\cmidrule(lr){1-9}
\textbf{CoM} & \textbf{81.2} & \textbf{88.7} & \textbf{90.9} & \textbf{92.6} & \textbf{81.2} & \textbf{87.1} & \textbf{90.9} & \underline{94.8} \\
\midrule
\rowcolor{white}\multicolumn{9}{l}{\textit{ViT-L/14}} \\
\rowcolor{white}Zero-shot & 71.4 & 71.4 & 71.4 & 71.4 & 72.0 & 81.0 & 84.0 & 92.4 \\
\cmidrule(lr){1-9}
Task Arithmetic & 74.1 & 74.4 & 74.4 & 74.8 & 76.3 & 83.6 & 86.1 & 93.4 \\
TIES & 74.6 & 75.2 & 75.0 & 75.5 & 77.3 & 84.2 & 86.5 & 93.7 \\
DARE-TIES & 74.1 & 74.5 & 74.5 & 74.9 & 75.4 & 83.6 & 85.4 & 93.5 \\
LiNeS & 73.9 & 74.4 & 74.4 & 74.8 & 75.3 & 83.6 & 86.0 & 93.5 \\
Consensus & 72.4 & 74.4 & 74.4 & 74.8 & 76.7 & 83.6 & 85.8 & 92.9 \\
TSV & 76.0 & 77.8 & 77.8 & 77.8 & 77.8 & 86.6 & 88.3 & 94.6 \\
KnOTS-TIES & 75.6 & 78.3 & 79.1 & 79.1 & 78.8 & 86.8 & 88.6 & 94.4 \\
Iso-C & 78.2 & 84.3 & 84.7 & 84.7 & 78.4 & 88.6 & 90.0 & 94.2 \\
Core\textsubscript{\tiny TSV} & 76.7 & 77.2 & 77.7 & 78.2 & 78.0 & 87.5 & 88.6 & 93.9 \\
Core\textsubscript{\tiny ISO} & \underline{86.0} & \underline{85.6} & \underline{85.6} & \underline{86.1} & 81.9 & \textbf{91.3} & \textbf{92.9} & \underline{95.9} \\
RegMean & 81.3 & 81.0 & 81.0 & 80.7 & \underline{83.5} & \underline{90.4} & \underline{92.2} & \textbf{96.2} \\
\cmidrule(lr){1-9}
\textbf{CoM} & \textbf{87.9} & \textbf{90.6} & \textbf{91.0} & \textbf{91.3} & \textbf{87.9} & 90.3 & 91.9 & 95.6 \\
\midrule
\rowcolor{white}\multicolumn{9}{l}{\textit{Llama-3.2-1B}} \\
\rowcolor{white}Zero-shot & -- & -- & -- & -- & 51.6 & 65.1 & 72.0 & 89.3 \\
\cmidrule(lr){1-9}
Task Arithmetic & 70.3 & 81.1 & 81.1 & 81.1 & 69.5 & 85.0 & 87.5 & 94.7 \\
TIES & 73.7 & 80.7 & 80.9 & 81.5 & 75.7 & 77.2 & 86.6 & 93.7 \\
DARE-TIES & 80.8 & 80.3 & 81.4 & 81.2 & 75.8 & 84.5 & 87.8 & 96.0 \\
LiNeS & 71.7 & 81.1 & 81.1 & 81.1 & 70.9 & 85.0 & 87.6 & 94.4 \\
Consensus & 71.5 & 71.2 & 79.5 & 79.7 & 70.5 & 80.8 & 88.3 & 95.3 \\
TSV & 74.9 & 87.5 & 87.5 & 87.5 & 74.0 & 87.4 & 90.5 & 95.3 \\
KnOTS-TIES & 66.7 & 82.1 & 81.2 & 82.4 & 66.8 & 86.6 & 88.4 & 95.2 \\
Iso-C & 53.4 & 53.4 & 53.5 & 72.9 & 52.0 & 66.7 & 75.3 & 94.8 \\
Core\textsubscript{\tiny TSV} & 75.5 & 88.0 & 89.5 & 89.5 & 73.9 & 90.0 & 89.8 & 96.2 \\
Core\textsubscript{\tiny ISO} & 54.6 & 64.2 & 83.3 & 83.2 & 54.7 & 80.4 & \underline{94.2} & 94.6 \\
RegMean & \underline{88.8} & \underline{92.5} & \underline{93.2} & \underline{93.3} & \underline{88.8} & \underline{92.8} & 93.5 & \underline{96.3} \\
\cmidrule(lr){1-9}
\textbf{CoM} & \textbf{90.9} & \textbf{95.3} & \textbf{96.4} & \textbf{96.3} & \textbf{90.9} & \textbf{95.3} & \textbf{95.2} & \textbf{96.4} \\
\bottomrule
\end{tabular}
\endgroup
\end{table}

Two regimes emerge from \cref{tab:merge_vs_ft}. With few examples per task, fine-tuning is unreliable, to the point that it is often not even preferred to the merged model on the held-out examples. In this regime, merging is the most effective way to exploit the available data, and CoM, which does not need labels, yields the strongest model. As the number of labeled examples grows, fine-tuning becomes effective and all methods converge to similar accuracies.

On Llama-3.2-1B, the CoM merge alone is on par with the best fine-tuned models even with $500$ examples per task. At that point, the question is no longer which merging method to use, but whether merging is needed at all, as there is enough labeled data to train on all tasks jointly; even then, starting from a merged model remains preferable to fine-tuning the pretrained one. Merging is thus most valuable when little or no labeled data is available, where CoM makes the best use of it.

Finally, this comparison only matches the amount of data, and does not account for the much higher computational cost and memory footprint of fine-tuning: while CoM requires a single forward pass per model, fine-tuning requires backpropagation through the whole backbone, optimizer states, and repeated runs to select the learning rate.

\clearpage
\section{Additional Results}
\label{app:group_results}
\subsection{Accuracies on each dataset}
\label{app:more_accuracies}
Following~\cite{stoica2025model,panariello2025accurate}, we report the normalized accuracies on each dataset in~\cref{tab:all_data_vitb32} for ViT-B/32, in~\cref{tab:all_data_vitl14} for ViT-L/14 and in~\cref{tab:all_data_llama} for Llama-3-8B with LoRA fine-tuning.
\input{tables/all_datasets_vision_b32}
\input{tables/all_datasets_vision_l14}
\input{tables/all_datasets_llama}

\subsection{Full Fine-Tuning on Language Tasks}
\label{app:fft_language}
For memory reasons, fully fine-tuning and merging the 8B model was not feasible in our setup, so we use a smaller Llama (Llama-3.2-1B) instead of Llama-3-8B. We fully fine-tune all parameters of the model on each of the six tasks with AdamW, a cosine schedule and an effective batch size of $32$; for each task, the learning rate is chosen among $\{5\!\times\!10^{-6}, 10^{-5}, 2\!\times\!10^{-5}\}$ and the number of epochs (at most three) by early stopping on the validation split. As for the other experiments, the hyperparameters of all baselines are tuned on the validation split, RegMean is used in its original setting, and CoM uses $500$ class-balanced samples per task. Results are reported in \cref{tab:fft_language,tab:fft_language_pertask}. CoM remains the best-performing method by a large margin, outperforming the strongest competitor, Iso-C, by $4.1$ points of normalized accuracy; RegMean, whose raw-token Gram matrices are dominated by the largest tasks, collapses on SICK. Core\textsubscript{\tiny TSV}, Core\textsubscript{\tiny ISO} and KnOTS are omitted, as they are designed for LoRA updates and are therefore not applicable to full fine-tuning.

\begin{table}[ht]
\centering
\caption{Absolute and normalized average accuracy (\%) of merged fully fine-tuned Llama-3.2-1B models on the six language tasks. CoM: mean over three calibration seeds ($\pm$ std); DARE-TIES: mean over three drop masks; other rows are deterministic.}
\label{tab:fft_language}
\rowcolors{4}{}{lightgray}
\begin{tabular}{lcc}
\toprule
Method & Abs. Acc. & Norm. Acc. \\
\midrule
Zero-shot & 45.47 & 50.86 \\
Individual FT & 89.56 & 100.00 \\
\midrule
Task Arithmetic & 77.64 & 86.85 \\
TIES & 77.31 & 86.39 \\
DARE-TIES & 77.13 & 86.29 \\
Consensus & 77.96 & 87.24 \\
LiNeS & 77.61 & 86.81 \\
RegMean & 79.49 & 88.82 \\
TSV & 79.38 & 88.70 \\
Iso-C & \underline{82.21} & \underline{91.96} \\
\midrule
\textbf{CoM} & \textbf{85.99} {\scriptsize $\pm$ 0.28} & \textbf{96.09} {\scriptsize $\pm$ 0.34} \\
\bottomrule
\end{tabular}
\end{table}
\vspace{2em}

\begin{table}[ht]
\centering
\setlength{\tabcolsep}{0.95em}
\caption{Per-task normalized test accuracy (\%) of merged fully fine-tuned Llama-3.2-1B models.}
\label{tab:fft_language_pertask}
\rowcolors{3}{}{lightgray}
\begin{tabular}{lcccccc|c}
\toprule
Method & SNLI & MNLI & SICK & QNLI & RTE & SciTail & Avg \\
\midrule
Zero-shot & 35.32 & 36.18 & 60.42 & 51.73 & 59.09 & 62.41 & 50.86 \\
\midrule
Task Arithmetic & 94.72 & 94.27 & 79.70 & 69.54 & 92.73 & 90.15 & 86.85 \\
TIES & 98.62 & 99.02 & 68.05 & 75.02 & 89.09 & 88.51 & 86.39 \\
DARE-TIES & 96.17 & 95.83 & 74.60 & 68.09 & 93.03 & 90.04 & 86.29 \\
Consensus & 93.07 & 92.02 & 85.38 & 68.90 & 93.64 & 90.45 & 87.24 \\
LiNeS & 95.53 & 95.51 & 76.87 & 69.83 & 92.73 & 90.40 & 86.81 \\
RegMean & 100.23 & 100.23 & 63.44 & 87.78 & 93.64 & 87.62 & 88.82 \\
TSV & 98.54 & 98.33 & 73.17 & 79.67 & 91.82 & 90.65 & 88.70 \\
Iso-C & 92.40 & 93.77 & 92.76 & 81.44 & 98.18 & 93.24 & 91.96 \\
\midrule
\textbf{CoM} & 98.40 & 97.18 & 92.40 & 97.50 & 99.09 & 91.99 & \textbf{96.09} \\
\bottomrule
\end{tabular}%
\end{table}

%% file: tables/variant_components_ablation.tex
\setlength{\intextsep}{0pt}

\begin{wraptable}[17]{r}{0.56\linewidth}
\vspace{-0.14em}
\centering
\small

\caption{Impact of individual components of CoM. \textit{MCS} denotes solving Merging Covariate Shift, \textit{Norm} refers to activation normalization, and \textit{FC} indicates feature correlation weighting. Values are averaged over three seeds. First row is RegMean with CoM's calibration set.}
\label{tab:variant_components_ablation}

\vspace{-0.25em}

\begingroup
\setlength{\tabcolsep}{3.5pt}
\renewcommand{\arraystretch}{1.0}
\setlength{\extrarowheight}{1pt}
\rowcolors{5}{}{lightgray}

\begin{tabularx}{\linewidth}{
    >{\centering\arraybackslash\hsize=0.72\hsize}X
    >{\centering\arraybackslash\hsize=0.72\hsize}X
    >{\centering\arraybackslash\hsize=0.72\hsize}X
    >{\centering\arraybackslash\hsize=1.28\hsize}X
    >{\centering\arraybackslash\hsize=1.28\hsize}X
    >{\centering\arraybackslash\hsize=1.28\hsize}X}
  \toprule
  \multicolumn{3}{c}{Components} & \multicolumn{3}{c}{Architecture} \\
  \cmidrule(lr){1-3} \cmidrule(lr){4-6}
  MCS & Norm & FC & ViT\mbox{-}B/32 & ViT\mbox{-}L/14 & Llama\mbox{-}3\mbox{-}8B \\
  \midrule
  \xmark & \xmark & \xmark & 72.77 & 80.69 & 97.31 \\
  \xmark & \cmark & \xmark & 74.32 & 83.05 & 97.52 \\
  \cmark & \xmark & \xmark & 83.45 & 84.26 & 99.71 \\
  \xmark & \xmark & \cmark & 82.79 & 85.66 & 95.02 \\
  \xmark & \cmark & \cmark & 84.02 & 88.27 & 97.52 \\
  \cmark & \cmark & \xmark & 85.83 & 87.38 & 99.76 \\
  \cmark & \xmark & \cmark & 91.03 & 90.12 & 95.03 \\
  \cmark & \cmark & \cmark & 92.74 & 91.35 & 99.63 \\
  \bottomrule
\end{tabularx}
\endgroup

\vspace{-0.8em}
\end{wraptable}

%% file: algorithms/chain_of_merges.tex
\setlength{\intextsep}{0pt}
\begin{algorithm}[b!]
\caption{--~ Chain of Merges}\label{alg:chain_merges}
\begin{algorithmic}[1]
\Require Model weights $\{\mW_i^l\}_{i=1,2, \dots, N}^{l=1,2, \dots, L}$, first layer inputs $\{\mX_i^1\}_{i=1}^N$, depth $L$
\vspace{0.5em}
\For{$l = 1$ to $L$}
    \For{$i = 1$ to $N$}
        \If{$l=1$}
            \State $\widehat{\mX}_i^1 \gets \mX_i^1$
        \Else
            \State $\widehat{\mX}_i^l \gets \boldsymbol{\sigma}(\mW_M^{l-1}\widehat{\mX}_i^{l-1})$
        \EndIf
        \State $\tilde{\mX}_i^l \gets \widehat{\mX}_i^l$ with each column (token) $\hat{\vx}_t$ scaled to $\hat{\vx}_t / \|\hat{\vx}_t\|_2$
        \State $\mG_i^{\,l} \gets \tfrac{1}{T_i}\tilde{\mX}_{i}^{\,l}\tilde{\mX}_{i}^{\,l\Ttop}$
        \State $\omega_i^{\,l} \gets \sum_{\,p\,<\,q} \big|(\mG^{\,l}_i)_{pq}\big|$
    \EndFor
    \State Update $\displaystyle \mW_M^l$ using \cref{eq:cm_weighted}
\EndFor
\State \Return $\{\mW_M^l\}_{l=1}^L$
\end{algorithmic}
\end{algorithm}

%% file: tables/samples_ablation_full.tex
\begin{table}[ht]
  \centering
  \caption{Average normalized accuracy of CoM (LoRA) varying the number of examples per task: numerical values of \cref{fig:samples_ablation}.}
  \label{tab:samples_ablation_full}
  \begingroup
  \setlength{\tabcolsep}{1.1em}
\renewcommand{\arraystretch}{0.92}
\setlength{\extrarowheight}{1pt}
\rowcolors{3}{lightgray}{}

\begin{tabular}{p{1.75cm}
                >{\centering\arraybackslash}p{0.64cm}
                >{\centering\arraybackslash}p{0.64cm}
                >{\centering\arraybackslash}p{0.64cm}
                >{\centering\arraybackslash}p{0.64cm}
                >{\centering\arraybackslash}p{0.64cm}
                >{\centering\arraybackslash}p{0.64cm}
                >{\centering\arraybackslash}p{0.64cm}
                >{\centering\arraybackslash}p{0.64cm}}
    \toprule
    \# of samples & 5 & 10 & 50 & 100 & 200 & 300 & 400 & 500 \\
    \midrule
    {ViT-B/32}
    & 76.94 & 81.95 & 88.74 & 91.02 & 92.24 & 92.57 & 92.67 & 92.74 \\
    {ViT-L/14}
    & 86.12 & 88.43 & 90.85 & 91.24 & 91.34 & 91.32 & 91.36 & 91.35 \\
    {Llama-3-8B}
    & 97.17 & 97.59 & 98.97 & 99.43 & 99.71 & 99.48 & 99.66 & 99.76 \\
    \bottomrule
  \end{tabular}
\endgroup
\end{table}

%% file: tables/all_datasets_vision_b32.tex
\begin{table}[!b]
\centering
\caption{Normalized accuracies of merged models on vision tasks with ViT-B/32, LoRA fine-tuning. Best results in bold, second-best underlined.}
\label{tab:all_data_vitb32}

\begingroup
\setlength{\tabcolsep}{5pt}
\renewcommand{\arraystretch}{1.0}
\setlength{\extrarowheight}{1pt}
\rowcolors{2}{}{lightgray}

\begin{tabular}{p{2.1cm}
                >{\centering\arraybackslash}p{0.9cm}
                >{\centering\arraybackslash}p{0.9cm}
                >{\centering\arraybackslash}p{0.9cm}
                >{\centering\arraybackslash}p{0.9cm}
                >{\centering\arraybackslash}p{0.9cm}
                >{\centering\arraybackslash}p{0.9cm}
                >{\centering\arraybackslash}p{0.9cm}
                >{\centering\arraybackslash}p{0.92cm}
                |>{\centering\arraybackslash}p{1.1cm}}
\toprule
\textbf{ViT-B/32} & Cars & DTD & ESAT & \makebox[0pt][c]{GTSRB} &
\makebox[0pt][c]{MNIST} & \makebox[0pt][c]{RESISC} & SUN & SVHN & \makebox[0pt][c]{\textbf{Average}} \\
\midrule
TA
& 81.97 & 73.99 & 49.01 & 41.93 & 53.45 & 71.57 & 97.47 & 42.08 & 63.93 \\
TIES
& 81.57 & 72.90 & 49.20 & 37.87 & 57.40 & 70.05 & 96.64 & 46.87 & 64.06 \\
DARE\textsubscript{\tiny{TIES}}
& 82.70 & 72.81 & 49.49 & 37.68 & 57.73 & 69.87 & 96.87 & 45.92 & 64.13 \\
Consensus TA
& 82.72 & 73.08 & 48.75 & 40.79 & 60.51 & 71.46 & 96.95 & 47.78 & 65.26 \\
LiNeS
& 81.59 & 74.91 & 50.84 & 39.88 & 52.81 & 71.07 & 97.99 & 39.27 & 63.54 \\
FisherAVG
& \underline{84.91} & 73.99 & \underline{81.74} & 53.19 & 51.15 & 69.58 & 96.42 & 34.05 & 68.13 \\
RegMean
& 77.20 & 64.14 & 36.25 & 50.47 & \underline{97.47} & 77.03 & 88.99 & \underline{94.99} & 73.32 \\
MaTS
& 84.47 & 73.72 & 80.10 & 52.00 & 53.10 & 70.15 & 96.77 & 36.36 & 68.33 \\
TSV
& 83.12 & 74.91 & 49.12 & 44.85 & 61.30 & 73.10 & 95.74 & 52.22 & 66.80 \\
Iso-C
& 79.07 & 82.84 & 52.45 & 75.62 & 68.93 & 79.11 & \textbf{98.70} & 50.49 & 73.40 \\
KnOTS\textsubscript{\tiny{TIES}}
& 82.87 & 73.54 & 48.11 & 48.67 & 71.44 & 70.87 & 95.52 & 55.51 & 68.32 \\
Core\textsubscript{\tiny{TSV}}
& 83.77 & 75.45 & 52.49 & 45.21 & 59.30 & 72.99 & 97.71 & 46.17 & 66.64 \\
Core\textsubscript{\tiny{ISO}}
& 83.29 & \underline{84.30} & 48.67 & \underline{82.43} & 73.02 & \underline{83.28} & \underline{98.65} & 55.37 & \underline{76.13} \\
\textbf{\methname (ours)}
& \textbf{90.58} & \textbf{87.80} & \textbf{89.99} & \textbf{91.99} & \textbf{99.42} & \textbf{94.54} & 89.52 & \textbf{98.11} & \textbf{92.74} \\
\bottomrule
\end{tabular}
\endgroup

\end{table}

%% file: tables/all_datasets_vision_l14.tex
\begin{table}[!b]
\centering
\caption{Normalized accuracies of merged models on vision tasks with ViT-L/14, LoRA fine-tuning. Best results in bold, second-best underlined.}
\label{tab:all_data_vitl14}

\begingroup
\setlength{\tabcolsep}{5pt}
\renewcommand{\arraystretch}{1.0}
\setlength{\extrarowheight}{1pt}
\rowcolors{2}{}{lightgray}

\begin{tabular}{p{2.1cm}
                >{\centering\arraybackslash}p{0.9cm}
                >{\centering\arraybackslash}p{0.9cm}
                >{\centering\arraybackslash}p{0.9cm}
                >{\centering\arraybackslash}p{0.9cm}
                >{\centering\arraybackslash}p{0.9cm}
                >{\centering\arraybackslash}p{0.9cm}
                >{\centering\arraybackslash}p{0.9cm}
                >{\centering\arraybackslash}p{0.92cm}
                |>{\centering\arraybackslash}p{1.1cm}}
\toprule
\textbf{ViT-L/14} & Cars & DTD & ESAT & \makebox[0pt][c]{GTSRB} & \makebox[0pt][c]{MNIST} & \makebox[0pt][c]{RESISC} & SUN & SVHN & \makebox[0pt][c]{\textbf{Average}} \\
\midrule
{TA}
& 80.07 & 79.73 & 65.51 & 60.01 & 82.27 & 79.52 & 86.68 & 64.95 & 74.84 \\
{TIES}
& 79.54 & 78.21 & 64.27 & 61.18 & 84.16 & 79.48 & 87.36 & 70.19 & 75.55 \\
DARE\textsubscript{\tiny{TIES}}
& 79.95 & 78.89 & 63.22 & 60.11 & 81.66 & 79.91 & 87.50 & 68.77 & 75.00 \\
Consensus TA
& 81.88 & 81.32 & 68.60 & 64.55 & 85.04 & 81.56 & 86.90 & 63.72 & 76.69 \\
LiNeS
& 79.70 & 79.73 & 66.04 & 57.88 & 81.86 & 78.24 & 86.48 & 64.10 & 74.25 \\
FisherAVG
& 82.60 & 81.09 & 66.68 & \underline{87.66} & 75.32 & 79.70 & 85.44 & 61.41 & 77.49 \\
{RegMean}
& 77.56 & 78.59 & 65.74 & 80.93 & \underline{94.25} & 90.86 & 88.16 & \underline{89.05} & 83.14 \\
MaTS
& 82.08 & 81.17 & 66.79 & 85.65 & 75.80 & 79.96 & 85.55 & 62.60 & 77.45 \\
{TSV}
& 82.58 & 80.11 & 64.69 & 69.31 & 83.01 & 83.68 & 87.91 & 70.91 & 77.77 \\
{Iso-C}
& 87.86 & \underline{89.14} & \underline{81.03} & 81.36 & 90.24 & 89.40 & \underline{88.77} & 69.78 & 84.70 \\
KnOTS\textsubscript{\tiny{TIES}}
& 82.53 & 80.41 & 64.58 & 68.81 & 88.87 & 82.32 & 88.26 & 76.78 & 79.07 \\
Core\textsubscript{\tiny{TSV}}
& 82.00 & 79.88 & 67.02 & 66.81 & 87.78 & 82.24 & 87.63 & 72.04 & 78.17 \\
Core\textsubscript{\tiny{ISO}}
& \textbf{91.37} & \textbf{90.89} & 79.90 & 86.02 & 88.50 & \textbf{91.33} & \textbf{89.77} & 70.87 & \underline{86.08} \\
\textbf{\methname (ours)}
& \underline{88.97} & 88.51 & \textbf{87.82} & \textbf{93.26} & \textbf{99.38} & \underline{91.14} & 83.87 & \textbf{97.85} & \textbf{91.35} \\
\bottomrule
\end{tabular}
\endgroup

\end{table}

%% file: tables/all_datasets_llama.tex
\begin{table}[t]
  \centering
\caption{Normalized accuracies of merged models on language with Llama-3-8B. Best results are highlighted in bold, second-best underlined.}
\label{tab:all_data_llama}
  \begingroup
\setlength{\tabcolsep}{5pt}
\renewcommand{\arraystretch}{1.0}
\setlength{\extrarowheight}{1pt}
\rowcolors{2}{}{lightgray}

\begin{tabular}{p{2.15cm}
                >{\centering\arraybackslash}p{1.23cm}
                >{\centering\arraybackslash}p{1.23cm}
                >{\centering\arraybackslash}p{1.23cm}
                >{\centering\arraybackslash}p{1.23cm}
                >{\centering\arraybackslash}p{1.23cm}
                >{\centering\arraybackslash}p{1.4cm}
                |>{\centering\arraybackslash}p{1.4cm}}
    \toprule
    \textbf{Method} & SNLI & MNLI & SICK & QNLI & RTE & SCITAIL & \textbf{Average} \\
    \midrule
{TA}
& 93.66 & 95.11 & 88.14 & 68.66 & 99.19 & 96.83 & 90.27 \\
{TIES}
& 95.07 & 96.32 & 83.46 & 70.70 & 100.00 & 96.35 & 90.32 \\
DARE\textsubscript{\tiny{TIES}}
& 94.84 & 96.40 & 81.62 & 71.66 & 99.19 & 96.07 & 89.96 \\
Consensus TA
& 94.26 & 94.63 & 88.78 & 66.43 & \underline{101.61} & 96.93 & 90.44 \\
LiNeS
& 92.87 & 95.35 & 89.14 & 73.24 & 97.58 & 96.64 & 90.80 \\
{RegMean}
& \textbf{99.91} & \textbf{98.65} & 82.02 & \underline{98.55} & 94.35 & 95.03 & \underline{94.75} \\
{TSV}
& 94.67 & 95.71 & 86.07 & 78.54 & 100.81 & 97.13 & 92.15 \\
{Iso-C}
& 84.52 & 51.50 & 93.14 & 53.94 & 66.94 & 75.05 & 70.85 \\
KnOTS\textsubscript{TIES}
& 89.48 & 93.64 & \underline{93.92} & 81.16 & \underline{101.61} & 97.71 & 92.92 \\
Core\textsubscript{TSV}
& 95.91 & 95.71 & 89.32 & 83.92 & \textbf{102.42} & \underline{97.86} & 94.19 \\
Core\textsubscript{ISO}
& 90.93 & 91.58 & 85.33 & 77.11 & 100.00 & 97.51 & 90.41 \\
\textbf{\methname (ours)}
& \underline{99.60} & \underline{97.64} & \textbf{100.28} & \textbf{99.98} & 101.34 & \textbf{99.72} & \textbf{99.76} \\
\bottomrule
  \end{tabular}
\endgroup
\end{table}

%% file: main.bbl
\begin{thebibliography}{71}
\providecommand{\natexlab}[1]{#1}
\providecommand{\url}[1]{\texttt{#1}}
\expandafter\ifx\csname urlstyle\endcsname\relax
  \providecommand{\doi}[1]{doi: #1}\else
  \providecommand{\doi}{doi: \begingroup \urlstyle{rm}\Url}\fi

\bibitem[Aghajanyan et~al.(2021)Aghajanyan, Gupta, and
  Zettlemoyer]{aghajanyan-etal-2021-intrinsic}
Armen Aghajanyan, Sonal Gupta, and Luke Zettlemoyer.
\newblock Intrinsic dimensionality explains the effectiveness of language model
  fine-tuning.
\newblock In Chengqing Zong, Fei Xia, Wenjie Li, and Roberto Navigli, editors,
  \emph{Association for Computational Linguistics}, pages 7319--7328, Online,
  August 2021. Association for Computational Linguistics.
\newblock \doi{10.18653/v1/2021.acl-long.568}.
\newblock URL \url{https://aclanthology.org/2021.acl-long.568/}.

\bibitem[Ainsworth et~al.(2023)Ainsworth, Hayase, and
  Srinivasa]{ainsworth2023git}
Samuel Ainsworth, Jonathan Hayase, and Siddhartha Srinivasa.
\newblock Git re-basin: Merging models modulo permutation symmetries.
\newblock In \emph{International Conference on Learning Representations}, 2023.

\bibitem[Arefin et~al.(2024)Arefin, Subbaraj, Gontier, LeCun, Rish,
  Shwartz-Ziv, and Pal]{arefin2024seq}
Md~Rifat Arefin, Gopeshh Subbaraj, Nicolas Gontier, Yann LeCun, Irina Rish,
  Ravid Shwartz-Ziv, and Christopher Pal.
\newblock Seq-vcr: Preventing collapse in intermediate transformer
  representations for enhanced reasoning.
\newblock \emph{International Conference on Learning Representations}, 2024.

\bibitem[Arpit et~al.(2016)Arpit, Zhou, Kota, and
  Govindaraju]{arpit2016normalization}
Devansh Arpit, Yingbo Zhou, Bhargava Kota, and Venu Govindaraju.
\newblock Normalization propagation: A parametric technique for removing
  internal covariate shift in deep networks.
\newblock In \emph{International Conference on Machine Learning}, pages
  1168--1176. PMLR, 2016.

\bibitem[Ba et~al.(2016)Ba, Kiros, and Hinton]{ba2016layer}
Jimmy~Lei Ba, Jamie~Ryan Kiros, and Geoffrey~E Hinton.
\newblock Layer normalization.
\newblock \emph{arXiv preprint arXiv:1607.06450}, 2016.

\bibitem[Barbero et~al.(2024)Barbero, Banino, Kapturowski, Kumaran, Ara{\'u}jo,
  Vitvitskyi, Pascanu, and Velickovic]{barbero2406transformers}
Federico Barbero, Andrea Banino, Steven Kapturowski, Dharshan Kumaran, Joao~GM
  Ara{\'u}jo, Alex Vitvitskyi, Razvan Pascanu, and Petar Velickovic.
\newblock Transformers need glasses! information over-squashing in language
  tasks.
\newblock In \emph{Advances in Neural Information Processing Systems}, 2024.

\bibitem[Bowman et~al.(2015)Bowman, Angeli, Potts, and Manning]{bowman2015snli}
Samuel~R Bowman, Gabor Angeli, Christopher Potts, and Christopher~D Manning.
\newblock A large annotated corpus for learning natural language inference.
\newblock In \emph{Conference on Empirical Methods in Natural Language
  Processing}, 2015.

\bibitem[Cheng et~al.(2017)Cheng, Han, and Lu]{cheng2017remote}
Gong Cheng, Junwei Han, and Xiaoqiang Lu.
\newblock Remote sensing image scene classification: Benchmark and state of the
  art.
\newblock \emph{Proceedings of the IEEE}, 105\penalty0 (10):\penalty0
  1865--1883, 2017.

\bibitem[Cimpoi et~al.(2014)Cimpoi, Maji, Kokkinos, Mohamed, and
  Vedaldi]{cimpoi2014describing}
Mircea Cimpoi, Subhransu Maji, Iasonas Kokkinos, Sammy Mohamed, and Andrea
  Vedaldi.
\newblock Describing textures in the wild.
\newblock In \emph{Proceedings of the IEEE Conference on Computer Vision and
  Pattern Recognition}, pages 3606--3613, 2014.

\bibitem[Cogswell et~al.(2016)Cogswell, Ahmed, Girshick, Zitnick, and
  Batra]{cogswell2015reducing}
Michael Cogswell, Faruk Ahmed, Ross Girshick, Larry Zitnick, and Dhruv Batra.
\newblock Reducing overfitting in deep networks by decorrelating
  representations.
\newblock In \emph{International Conference on Learning Representations}, 2016.

\bibitem[Daheim et~al.(2024)Daheim, M{\"o}llenhoff, Ponti, Gurevych, and
  Khan]{daheim2024model}
Nico Daheim, Thomas M{\"o}llenhoff, Edoardo~Maria Ponti, Iryna Gurevych, and
  Mohammad~Emtiyaz Khan.
\newblock Model merging by uncertainty-based gradient matching.
\newblock \emph{International Conference on Learning Representations}, 2024.

\bibitem[Dosovitskiy et~al.(2021)Dosovitskiy, Beyer, Kolesnikov, Weissenborn,
  Zhai, Unterthiner, Dehghani, Minderer, Heigold, Gelly,
  et~al.]{dosovitskiyimage}
Alexey Dosovitskiy, Lucas Beyer, Alexander Kolesnikov, Dirk Weissenborn,
  Xiaohua Zhai, Thomas Unterthiner, Mostafa Dehghani, Matthias Minderer, Georg
  Heigold, Sylvain Gelly, et~al.
\newblock An image is worth 16x16 words: Transformers for image recognition at
  scale.
\newblock In \emph{International Conference on Learning Representations}, 2021.

\bibitem[Dowson and Landau(1982)]{dowson1982frechet}
DC~Dowson and BV~Landau.
\newblock The fr{\'e}chet distance between multivariate normal distributions.
\newblock \emph{Journal of multivariate analysis}, 12\penalty0 (3):\penalty0
  450--455, 1982.

\bibitem[Gargiulo et~al.(2025)Gargiulo, Crisostomi, Bucarelli, Scardapane,
  Silvestri, and Rodola]{gargiulo2025task}
Antonio~Andrea Gargiulo, Donato Crisostomi, Maria~Sofia Bucarelli, Simone
  Scardapane, Fabrizio Silvestri, and Emanuele Rodola.
\newblock Task singular vectors: Reducing task interference in model merging.
\newblock In \emph{Proceedings of the IEEE Conference on Computer Vision and
  Pattern Recognition}, pages 18695--18705, 2025.

\bibitem[Grattafiori et~al.(2024)Grattafiori, Dubey, Jauhri, Pandey, Kadian,
  Al-Dahle, Letman, Mathur, Schelten, Vaughan, et~al.]{grattafiori2024llama}
Aaron Grattafiori, Abhimanyu Dubey, Abhinav Jauhri, Abhinav Pandey, Abhishek
  Kadian, Ahmad Al-Dahle, Aiesha Letman, Akhil Mathur, Alan Schelten, Alex
  Vaughan, et~al.
\newblock The llama 3 herd of models.
\newblock \emph{arXiv preprint arXiv:2407.21783}, 2024.

\bibitem[He et~al.(2024)He, Hu, Lin, Zhang, and Zhao]{he2024localize}
Yifei He, Yuzheng Hu, Yong Lin, Tong Zhang, and Han Zhao.
\newblock Localize-and-stitch: Efficient model merging via sparse task
  arithmetic.
\newblock \emph{Transactions on Machine Learning Research}, 2024.

\bibitem[Helber et~al.(2019)Helber, Bischke, Dengel, and
  Borth]{helber2019eurosat}
Patrick Helber, Benjamin Bischke, Andreas Dengel, and Damian Borth.
\newblock Eurosat: A novel dataset and deep learning benchmark for land use and
  land cover classification.
\newblock \emph{IEEE Journal of Selected Topics in Applied Earth Observations
  and Remote Sensing}, 12\penalty0 (7):\penalty0 2217--2226, 2019.

\bibitem[Hoerl and Kennard(1970)]{hoerl1970ridge}
Arthur~E Hoerl and Robert~W Kennard.
\newblock Ridge regression: Biased estimation for nonorthogonal problems.
\newblock \emph{Technometrics}, 12\penalty0 (1):\penalty0 55--67, 1970.

\bibitem[Hu et~al.(2022)Hu, yelong shen, Wallis, Allen-Zhu, Li, Wang, Wang, and
  Chen]{hu2022lora}
Edward~J Hu, yelong shen, Phillip Wallis, Zeyuan Allen-Zhu, Yuanzhi Li, Shean
  Wang, Lu~Wang, and Weizhu Chen.
\newblock Lo{RA}: Low-rank adaptation of large language models.
\newblock In \emph{International Conference on Learning Representations}, 2022.

\bibitem[Huang et~al.(2024)Huang, Ye, Chen, He, Yue, and Ouyang]{huang2024emr}
Chenyu Huang, Peng Ye, Tao Chen, Tong He, Xiangyu Yue, and Wanli Ouyang.
\newblock Emr-merging: Tuning-free high-performance model merging.
\newblock \emph{Advances in Neural Information Processing Systems},
  37:\penalty0 122741--122769, 2024.

\bibitem[Huang and Yu(2020)]{huang2020internal}
You Huang and Yuanlong Yu.
\newblock An internal covariate shift bounding algorithm for deep neural
  networks by unitizing layers' outputs.
\newblock In \emph{Proceedings of the IEEE Conference on Computer Vision and
  Pattern Recognition}, pages 8465--8473, 2020.

\bibitem[Ilharco et~al.(2022)Ilharco, Wortsman, Gadre, Song, Hajishirzi,
  Kornblith, Farhadi, and Schmidt]{ilharco2022patching}
Gabriel Ilharco, Mitchell Wortsman, Samir~Yitzhak Gadre, Shuran Song, Hannaneh
  Hajishirzi, Simon Kornblith, Ali Farhadi, and Ludwig Schmidt.
\newblock Patching open-vocabulary models by interpolating weights.
\newblock \emph{Advances in Neural Information Processing Systems}, 2022.

\bibitem[Ilharco et~al.(2023)Ilharco, Ribeiro, Wortsman, Schmidt, Hajishirzi,
  and Farhadi]{ilharco2023editing}
Gabriel Ilharco, Marco~Tulio Ribeiro, Mitchell Wortsman, Ludwig Schmidt,
  Hannaneh Hajishirzi, and Ali Farhadi.
\newblock Editing models with task arithmetic.
\newblock In \emph{International Conference on Learning Representations}, 2023.

\bibitem[Ioffe and Szegedy(2015)]{ioffe2015batch}
Sergey Ioffe and Christian Szegedy.
\newblock Batch normalization: Accelerating deep network training by reducing
  internal covariate shift.
\newblock In \emph{International Conference on Machine Learning}, pages
  448--456. pmlr, 2015.

\bibitem[Jin et~al.(2023)Jin, Ren, Preotiuc-Pietro, and Cheng]{jin2023dataless}
Xisen Jin, Xiang Ren, Daniel Preotiuc-Pietro, and Pengxiang Cheng.
\newblock Dataless knowledge fusion by merging weights of language models.
\newblock In \emph{International Conference on Learning Representations}, 2023.

\bibitem[Jordan et~al.(2023)Jordan, Sedghi, Saukh, Entezari, and
  Neyshabur]{jordan2022repair}
Keller Jordan, Hanie Sedghi, Olga Saukh, Rahim Entezari, and Behnam Neyshabur.
\newblock {REPAIR}: Renormalizing permuted activations for interpolation
  repair.
\newblock In \emph{International Conference on Learning Representations}, 2023.

\bibitem[Kalamkar et~al.(2019)Kalamkar, Mudigere, Mellempudi, Das, Banerjee,
  Avancha, Vooturi, Jammalamadaka, Huang, Yuen, et~al.]{kalamkar2019study}
Dhiraj Kalamkar, Dheevatsa Mudigere, Naveen Mellempudi, Dipankar Das, Kunal
  Banerjee, Sasikanth Avancha, Dharma~Teja Vooturi, Nataraj Jammalamadaka,
  Jianyu Huang, Hector Yuen, et~al.
\newblock A study of bfloat16 for deep learning training.
\newblock \emph{arXiv preprint arXiv:1905.12322}, 2019.

\bibitem[Khot et~al.(2018)Khot, Sabharwal, and Clark]{khot2018scitail}
Tushar Khot, Ashish Sabharwal, and Peter Clark.
\newblock Scitail: A textual entailment dataset from science question
  answering.
\newblock In \emph{Proceedings of the AAAI Conference on Artificial
  Intelligence}, volume~32, 2018.

\bibitem[Krause et~al.(2013)Krause, Stark, Deng, and Fei-Fei]{krause20133d}
Jonathan Krause, Michael Stark, Jia Deng, and Li~Fei-Fei.
\newblock 3d object representations for fine-grained categorization.
\newblock In \emph{Proceedings of the IEEE international conference on computer
  vision workshops}, pages 554--561, 2013.

\bibitem[Kumar et~al.(2022)Kumar, Raghunathan, Jones, Ma, and
  Liang]{kumar2022fine}
Ananya Kumar, Aditi Raghunathan, Robbie Jones, Tengyu Ma, and Percy Liang.
\newblock Fine-tuning can distort pretrained features and underperform
  out-of-distribution.
\newblock \emph{International Conference on Learning Representations}, 2022.

\bibitem[LeCun et~al.(2002)LeCun, Bottou, Bengio, and
  Haffner]{lecun2002gradient}
Yann LeCun, L{\'e}on Bottou, Yoshua Bengio, and Patrick Haffner.
\newblock Gradient-based learning applied to document recognition.
\newblock \emph{Proceedings of the IEEE}, 86\penalty0 (11):\penalty0
  2278--2324, 2002.

\bibitem[Lee et~al.(2025)Lee, Liu, Wang, Wang, Cai, and Wu]{lee2025dynamic}
Sanwoo Lee, Jiahao Liu, Qifan Wang, Jingang Wang, Xunliang Cai, and Yunfang Wu.
\newblock Dynamic fisher-weighted model merging via {B}ayesian optimization.
\newblock In Luis Chiruzzo, Alan Ritter, and Lu~Wang, editors,
  \emph{Proceedings of the 2025 Conference of the Nations of the Americas
  Chapter of the Association for Computational Linguistics}, pages 4923--4935,
  Albuquerque, New Mexico, April 2025. Association for Computational
  Linguistics.
\newblock ISBN 979-8-89176-189-6.
\newblock \doi{10.18653/v1/2025.naacl-long.254}.
\newblock URL \url{https://aclanthology.org/2025.naacl-long.254/}.

\bibitem[Lu et~al.(2024)Lu, Fan, Wei, Qu, Chen, and Cheng]{lu2024twin}
Zhenyi Lu, Chenghao Fan, Wei Wei, Xiaoye Qu, Dangyang Chen, and Yu~Cheng.
\newblock Twin-merging: Dynamic integration of modular expertise in model
  merging.
\newblock \emph{Advances in Neural Information Processing Systems},
  37:\penalty0 78905--78935, 2024.

\bibitem[Marczak et~al.(2025)Marczak, Magistri, Cygert, Twardowski, Bagdanov,
  and van~de Weijer]{marczak2025no}
Daniel Marczak, Simone Magistri, Sebastian Cygert, Bart{\l}omiej Twardowski,
  Andrew~D Bagdanov, and Joost van~de Weijer.
\newblock No task left behind: Isotropic model merging with common and
  task-specific subspaces.
\newblock In \emph{International Conference on Machine Learning}, 2025.

\bibitem[Marelli et~al.(2014)Marelli, Bentivogli, Baroni, Bernardi, Menini, and
  Zamparelli]{marelli2014semeval}
Marco Marelli, Luisa Bentivogli, Marco Baroni, Raffaella Bernardi, Stefano
  Menini, and Roberto Zamparelli.
\newblock Semeval-2014 task 1: Evaluation of compositional distributional
  semantic models on full sentences through semantic relatedness and textual
  entailment.
\newblock In \emph{Proceedings of the 8th international workshop on semantic
  evaluation (SemEval 2014)}, pages 1--8, 2014.

\bibitem[Matena and Raffel(2022)]{matena2022merging}
Michael~S Matena and Colin~A Raffel.
\newblock Merging models with fisher-weighted averaging.
\newblock \emph{Advances in Neural Information Processing Systems}, 2022.

\bibitem[McMahan et~al.(2017)McMahan, Moore, Ramage, Hampson, and
  y~Arcas]{mcmahan2017communication}
Brendan McMahan, Eider Moore, Daniel Ramage, Seth Hampson, and Blaise~Aguera
  y~Arcas.
\newblock Communication-efficient learning of deep networks from decentralized
  data.
\newblock In \emph{International Conference on Artificial Intelligence and
  Statistics}. PMLR, 2017.

\bibitem[Morcos et~al.(2018)Morcos, Barrett, Rabinowitz, and
  Botvinick]{morcos2018importance}
Ari~S Morcos, David~GT Barrett, Neil~C Rabinowitz, and Matthew Botvinick.
\newblock On the importance of single directions for generalization.
\newblock In \emph{International Conference on Learning Representations}, 2018.

\bibitem[Netzer et~al.(2011)Netzer, Wang, Coates, Bissacco, Wu, Ng,
  et~al.]{netzer2011reading}
Yuval Netzer, Tao Wang, Adam Coates, Alessandro Bissacco, Baolin Wu, Andrew~Y
  Ng, et~al.
\newblock Reading digits in natural images with unsupervised feature learning.
\newblock In \emph{NIPS workshop on deep learning and unsupervised feature
  learning}, volume 2011, page~7. Granada, 2011.

\bibitem[Ortiz-Jimenez et~al.(2023)Ortiz-Jimenez, Favero, and
  Frossard]{ortizjimenez2023tangent}
Guillermo Ortiz-Jimenez, Alessandro Favero, and Pascal Frossard.
\newblock Task arithmetic in the tangent space: Improved editing of pre-trained
  models.
\newblock In \emph{Advances in Neural Information Processing Systems}, pages
  66727--66754, 2023.

\bibitem[Panariello et~al.(2025)Panariello, Marczak, Magistri, Porrello,
  Twardowski, Bagdanov, Calderara, and van~de Weijer]{panariello2025accurate}
Aniello Panariello, Daniel Marczak, Simone Magistri, Angelo Porrello,
  Bart{\l}omiej Twardowski, Andrew~D Bagdanov, Simone Calderara, and Joost
  van~de Weijer.
\newblock Accurate and efficient low-rank model merging in core space.
\newblock \emph{Advances in Neural Information Processing Systems}, 2025.

\bibitem[Pe{\~n}a et~al.(2023)Pe{\~n}a, Medeiros, Dubail, Aminbeidokhti,
  Granger, and Pedersoli]{pena2023rebasin}
Fidel A~Guerrero Pe{\~n}a, Heitor~Rapela Medeiros, Thomas Dubail, Masih
  Aminbeidokhti, Eric Granger, and Marco Pedersoli.
\newblock Re-basin via implicit {S}inkhorn differentiation.
\newblock In \emph{Proceedings of the IEEE Conference on Computer Vision and
  Pattern Recognition}, 2023.

\bibitem[Porrello et~al.(2026)Porrello, Buzzega, Dangel, Sommariva, Salami,
  Bonicelli, and Calderara]{porrello2026dataless}
Angelo Porrello, Pietro Buzzega, Felix Dangel, Thomas Sommariva, Riccardo
  Salami, Lorenzo Bonicelli, and Simone Calderara.
\newblock Dataless weight disentanglement in task arithmetic via
  kronecker-factored approximate curvature.
\newblock In \emph{International Conference on Learning Representations}, 2026.

\bibitem[Radford et~al.(2021)Radford, Kim, Hallacy, Ramesh, Goh, Agarwal,
  Sastry, Askell, Mishkin, Clark, et~al.]{radford2021learning}
Alec Radford, Jong~Wook Kim, Chris Hallacy, Aditya Ramesh, Gabriel Goh,
  Sandhini Agarwal, Girish Sastry, Amanda Askell, Pamela Mishkin, Jack Clark,
  et~al.
\newblock Learning transferable visual models from natural language
  supervision.
\newblock In \emph{International Conference on Machine Learning}, 2021.

\bibitem[Raffel et~al.(2020)Raffel, Shazeer, Roberts, Lee, Narang, Matena,
  Zhou, Li, and Liu]{raffel2020exploring}
Colin Raffel, Noam Shazeer, Adam Roberts, Katherine Lee, Sharan Narang, Michael
  Matena, Yanqi Zhou, Wei Li, and Peter~J Liu.
\newblock Exploring the limits of transfer learning with a unified text-to-text
  transformer.
\newblock \emph{Journal of Machine Learning Research}, 21\penalty0
  (140):\penalty0 1--67, 2020.

\bibitem[Rinaldi et~al.(2025)Rinaldi, Capitani, Bonicelli, Crisostomi, Bolelli,
  Ficarra, Rodol{\`a}, Calderara, and Porrello]{rinaldi2025update}
Filippo Rinaldi, Giacomo Capitani, Lorenzo Bonicelli, Donato Crisostomi,
  Federico Bolelli, Elisa Ficarra, Emanuele Rodol{\`a}, Simone Calderara, and
  Angelo Porrello.
\newblock Update your transformer to the latest release: Re-basin of task
  vectors.
\newblock In \emph{International Conference on Machine Learning}, 2025.

\bibitem[Rombach et~al.(2022)Rombach, Blattmann, Lorenz, Esser, and
  Ommer]{Rombach_2022_CVPR}
Robin Rombach, Andreas Blattmann, Dominik Lorenz, Patrick Esser, and Bj\"orn
  Ommer.
\newblock High-resolution image synthesis with latent diffusion models.
\newblock In \emph{Proceedings of the IEEE Conference on Computer Vision and
  Pattern Recognition}, pages 10684--10695, June 2022.

\bibitem[Salami et~al.(2025)Salami, Buzzega, Mosconi, Bonato, Sabetta, and
  Calderara]{salami2025closed}
Riccardo Salami, Pietro Buzzega, Matteo Mosconi, Jacopo Bonato, Luigi Sabetta,
  and Simone Calderara.
\newblock Closed-form merging of parameter-efficient modules for federated
  continual learning.
\newblock \emph{International Conference on Learning Representations}, 2025.

\bibitem[Singh and Jaggi(2020)]{singh2020model}
Sidak~Pal Singh and Martin Jaggi.
\newblock Model fusion via optimal transport.
\newblock \emph{Advances in Neural Information Processing Systems},
  33:\penalty0 22045--22055, 2020.

\bibitem[Stallkamp et~al.(2011)Stallkamp, Schlipsing, Salmen, and
  Igel]{stallkamp2011german}
Johannes Stallkamp, Marc Schlipsing, Jan Salmen, and Christian Igel.
\newblock The german traffic sign recognition benchmark: a multi-class
  classification competition.
\newblock In \emph{The 2011 international joint conference on neural networks},
  pages 1453--1460. IEEE, 2011.

\bibitem[Stoica et~al.(2024)Stoica, Bolya, Bjorner, Ramesh, Hearn, and
  Hoffman]{stoica2023zipit}
George Stoica, Daniel Bolya, Jakob Bjorner, Pratik Ramesh, Taylor Hearn, and
  Judy Hoffman.
\newblock {ZipIt!} merging models from different tasks without training.
\newblock In \emph{International Conference on Learning Representations}, 2024.

\bibitem[Stoica et~al.(2025)Stoica, Ramesh, Ecsedi, Choshen, and
  Hoffman]{stoica2025model}
George Stoica, Pratik Ramesh, Boglarka Ecsedi, Leshem Choshen, and Judy
  Hoffman.
\newblock Model merging with svd to tie the knots.
\newblock \emph{International Conference on Learning Representations}, 2025.

\bibitem[Tam et~al.(2024)Tam, Bansal, and Raffel]{tam2023merging}
Derek Tam, Mohit Bansal, and Colin Raffel.
\newblock Merging by matching models in task parameter subspaces.
\newblock \emph{Transactions on Machine Learning Research}, 2024.

\bibitem[Tang et~al.(2024)Tang, Shen, Luo, Yin, Zhang, and
  Tao]{tang2024merging}
Anke Tang, Li~Shen, Yong Luo, Nan Yin, Lefei Zhang, and Dacheng Tao.
\newblock Merging multi-task models via weight-ensembling mixture of experts.
\newblock \emph{International Conference on Machine Learning}, 2024.

\bibitem[Tatro et~al.(2020)Tatro, Chen, Das, Melnyk, Sattigeri, and
  Lai]{tatro2020optimizing}
Norman Tatro, Pin-Yu Chen, Payel Das, Igor Melnyk, Prasanna Sattigeri, and
  Rongjie Lai.
\newblock Optimizing mode connectivity via neuron alignment.
\newblock \emph{Advances in Neural Information Processing Systems},
  33:\penalty0 15300--15311, 2020.

\bibitem[Touvron et~al.(2023)Touvron, Lavril, Izacard, Martinet, Lachaux,
  Lacroix, Rozi{\`e}re, Goyal, Hambro, Azhar, et~al.]{touvron2023llama}
Hugo Touvron, Thibaut Lavril, Gautier Izacard, Xavier Martinet, Marie-Anne
  Lachaux, Timoth{\'e}e Lacroix, Baptiste Rozi{\`e}re, Naman Goyal, Eric
  Hambro, Faisal Azhar, et~al.
\newblock Llama: Open and efficient foundation language models.
\newblock \emph{arXiv:2302.13971}, 2023.

\bibitem[Villani(2008)]{villani2008optimal}
C{\'e}dric Villani.
\newblock \emph{Optimal transport: old and new}, volume 338.
\newblock Springer, 2008.

\bibitem[Wang et~al.(2019)Wang, Singh, Michael, Hill, Levy, and
  Bowman]{wang2018glue}
Alex Wang, Amanpreet Singh, Julian Michael, Felix Hill, Omer Levy, and Samuel
  Bowman.
\newblock Glue: A multi-task benchmark and analysis platform for natural
  language understanding.
\newblock In \emph{International Conference on Learning Representations}, 2019.

\bibitem[Wang et~al.(2024)Wang, Dimitriadis, Ortiz-Jimenez, Fleuret, and
  Frossard]{wang2024localizing}
Ke~Wang, Nikolaos Dimitriadis, Guillermo Ortiz-Jimenez, Fran{\c{c}}ois Fleuret,
  and Pascal Frossard.
\newblock Localizing task information for improved model merging and
  compression.
\newblock \emph{International Conference on Machine Learning}, 2024.

\bibitem[Wang et~al.(2025)Wang, Dimitriadis, Favero, Ortiz-Jimenez, Fleuret,
  and Frossard]{wang2025lines}
Ke~Wang, Nikolaos Dimitriadis, Alessandro Favero, Guillermo Ortiz-Jimenez,
  Francois Fleuret, and Pascal Frossard.
\newblock Lines: Post-training layer scaling prevents forgetting and enhances
  model merging.
\newblock \emph{International Conference on Learning Representations}, 2025.

\bibitem[Williams et~al.(2018)Williams, Nangia, and Bowman]{williams2017broad}
Adina Williams, Nikita Nangia, and Samuel~R Bowman.
\newblock A broad-coverage challenge corpus for sentence understanding through
  inference.
\newblock In \emph{Proceedings of the Conference of the North American Chapter
  of the Association for Computational Linguistics}, 2018.

\bibitem[Wortsman et~al.(2022{\natexlab{a}})Wortsman, Ilharco, Gadre, Roelofs,
  Gontijo-Lopes, Morcos, Namkoong, Farhadi, Carmon, Kornblith,
  et~al.]{wortsman2022model}
Mitchell Wortsman, Gabriel Ilharco, Samir~Ya Gadre, Rebecca Roelofs, Raphael
  Gontijo-Lopes, Ari~S Morcos, Hongseok Namkoong, Ali Farhadi, Yair Carmon,
  Simon Kornblith, et~al.
\newblock Model soups: averaging weights of multiple fine-tuned models improves
  accuracy without increasing inference time.
\newblock In \emph{International Conference on Machine Learning},
  2022{\natexlab{a}}.

\bibitem[Wortsman et~al.(2022{\natexlab{b}})Wortsman, Ilharco, Kim, Li,
  Kornblith, Roelofs, Lopes, Hajishirzi, Farhadi, Namkoong,
  et~al.]{wortsman2022robust}
Mitchell Wortsman, Gabriel Ilharco, Jong~Wook Kim, Mike Li, Simon Kornblith,
  Rebecca Roelofs, Raphael~Gontijo Lopes, Hannaneh Hajishirzi, Ali Farhadi,
  Hongseok Namkoong, et~al.
\newblock Robust fine-tuning of zero-shot models.
\newblock In \emph{Proceedings of the IEEE Conference on Computer Vision and
  Pattern Recognition}, pages 7959--7971, 2022{\natexlab{b}}.

\bibitem[Wu et~al.(2024)Wu, Ajorlou, Wang, Jegelka, and Jadbabaie]{wu2024role}
Xinyi Wu, Amir Ajorlou, Yifei Wang, Stefanie Jegelka, and Ali Jadbabaie.
\newblock On the role of attention masks and layernorm in transformers.
\newblock \emph{Advances in Neural Information Processing Systems},
  37:\penalty0 14774--14809, 2024.

\bibitem[Xiao et~al.(2016)Xiao, Ehinger, Hays, Torralba, and
  Oliva]{xiao2016sun}
Jianxiong Xiao, Krista~A Ehinger, James Hays, Antonio Torralba, and Aude Oliva.
\newblock Sun database: Exploring a large collection of scene categories.
\newblock \emph{International Journal of Computer Vision}, 119\penalty0
  (1):\penalty0 3--22, 2016.

\bibitem[Xu et~al.(2025)Xu, Li, and Zhang]{xu2025scalable}
Jing Xu, Jiazheng Li, and Jingzhao Zhang.
\newblock Scalable model merging with progressive layer-wise distillation.
\newblock \emph{International Conference on Machine Learning}, 2025.

\bibitem[Yadav et~al.(2023)Yadav, Tam, Choshen, Raffel, and
  Bansal]{yadav2023ties}
Prateek Yadav, Derek Tam, Leshem Choshen, Colin~A Raffel, and Mohit Bansal.
\newblock Ties-merging: Resolving interference when merging models.
\newblock \emph{Advances in Neural Information Processing Systems},
  36:\penalty0 7093--7115, 2023.

\bibitem[Yang et~al.(2024)Yang, Wang, Shen, Liu, Guo, Wang, and
  Tao]{yang2023adamerging}
Enneng Yang, Zhenyi Wang, Li~Shen, Shiwei Liu, Guibing Guo, Xingwei Wang, and
  Dacheng Tao.
\newblock Adamerging: Adaptive model merging for multi-task learning.
\newblock \emph{International Conference on Learning Representations}, 2024.

\bibitem[Yoshida et~al.(2025)Yoshida, Naraki, Horie, Yamaki, Shimizu, Saito,
  McAuley, and Naganuma]{yoshida2025mastering}
Kotaro Yoshida, Yuji Naraki, Takafumi Horie, Ryosuke Yamaki, Ryotaro Shimizu,
  Yuki Saito, Julian McAuley, and Hiroki Naganuma.
\newblock Mastering task arithmetic: $\tau$jp as a key indicator for weight
  disentanglement.
\newblock In \emph{International Conference on Learning Representations}, 2025.

\bibitem[Yu et~al.(2024)Yu, Yu, Yu, Huang, and Li]{yu2024language}
Le~Yu, Bowen Yu, Haiyang Yu, Fei Huang, and Yongbin Li.
\newblock Language models are super mario: Absorbing abilities from homologous
  models as a free lunch.
\newblock In \emph{International Conference on Machine Learning}, 2024.

\bibitem[Zeng et~al.(2025)Zeng, Guo, Zhu, Shen, and Tang]{zeng2025robustmerge}
Fanhu Zeng, Haiyang Guo, Fei Zhu, Li~Shen, and Hao Tang.
\newblock {RobustMerge}: Parameter-efficient model merging for {MLLMs} with
  direction robustness.
\newblock In \emph{Advances in Neural Information Processing Systems}, 2025.

\end{thebibliography}
